\documentclass[letterpaper]{article} % DO NOT CHANGE THIS
\usepackage{aaai2026}  % DO NOT CHANGE THIS
\usepackage{times}  % DO NOT CHANGE THIS
\usepackage{helvet}  % DO NOT CHANGE THIS
\nocopyright % Remove copyright notice for arXiv version
\usepackage{courier}  % DO NOT CHANGE THIS
\usepackage[hyphens]{url}  % DO NOT CHANGE THIS
\usepackage{graphicx} % DO NOT CHANGE THIS
\usepackage{natbib}  % DO NOT CHANGE THIS AND DO NOT ADD ANY OPTIONS TO IT
\usepackage{caption} % DO NOT CHANGE THIS AND DO NOT ADD ANY OPTIONS TO IT
\usepackage{algorithm}
\usepackage{algpseudocode}
\usepackage{xcolor}
\usepackage{amsthm}
\usepackage{amsmath}
\usepackage{amssymb}
\usepackage{enumitem}
\usepackage{tabularray}
\usepackage{rotating}
\usepackage{soul}
\usepackage{subcaption}

\theoremstyle{plain}
\newtheorem{theorem}{Theorem}

\newtheorem{lemma}{Lemma}

\theoremstyle{definition}
\newtheorem{definition}{Definition}

\theoremstyle{remark}
\newtheorem{remark}{Remark}
\definecolor{FrenchPass}{rgb}{0.741,0.878,0.996}

\usepackage{newfloat}
\usepackage{listings}
\DeclareCaptionStyle{ruled}{labelfont=normalfont,labelsep=colon,strut=off} % DO NOT CHANGE THIS
\floatstyle{ruled}
\newfloat{listing}{tb}{lst}{}
\floatname{listing}{Listing}
\title{Beyond Binary Classification: A Semi-supervised Approach to Generalized AI-generated Image Detection}
\author{
    Hong-Hanh Nguyen-Le \textsuperscript{\rm 1},
    Van-Tuan Tran\textsuperscript{\rm 2},
    Dinh-Thuc Nguyen\textsuperscript{\rm 3},
    Nhien-An Le-Khac\textsuperscript{\rm 1}
}
\affiliations{
    \textsuperscript{\rm 1} School of Computer Science, University College Dublin, Ireland\\
    \textsuperscript{\rm 2} School of Computer Science and Statistics, Trinity College Dublin, Ireland\\
    \textsuperscript{\rm 3} Department of Knowledge Engineering, University of Science, VNU-HCMC, Vietnam

    hong-hanh.nguyen-le@ucdconnect.ie, tranva@tcd.ie, ndthuc@fit.hcmus.edu.vn, an.lekhac@ucd.ie
}

\begin{document}

\maketitle

\begin{abstract}
The rapid advancement of generators (e.g., StyleGAN, Midjourney, DALL-E) has produced highly realistic synthetic images, posing significant challenges to digital media authenticity. These generators are typically based on a few core architectural families, primarily Generative Adversarial Networks (GANs) and Diffusion Models (DMs). A critical vulnerability in current forensics is the failure of detectors to achieve cross-generator generalization, especially when crossing architectural boundaries (e.g., from GANs to DMs). We hypothesize that this gap stems from fundamental differences in the artifacts produced by these \textbf{distinct architectures}. In this work, we provide a theoretical analysis explaining how the distinct optimization objectives of the GAN and DM architectures lead to different manifold coverage behaviors. We demonstrate that GANs permit partial coverage, often leading to boundary artifacts, while DMs enforce complete coverage, resulting in over-smoothing patterns. Motivated by this analysis, we propose the \textbf{Tri}archy \textbf{Detect}or (TriDetect), a semi-supervised approach that enhances binary classification by discovering latent architectural patterns within the "fake" class. TriDetect employs balanced cluster assignment via the Sinkhorn-Knopp algorithm and a cross-view consistency mechanism, encouraging the model to learn fundamental architectural distincts. We evaluate our approach on two standard benchmarks and three in-the-wild datasets against 13 baselines to demonstrate its generalization capability to unseen generators.
\end{abstract}

% Uncomment the following to link to your code, datasets, an extended version or similar.
% You must keep this block between (not within) the abstract and the main body of the paper.
% \begin{links}
%     \link{Code}{https://aaai.org/example/code}
%     \link{Datasets}{https://aaai.org/example/datasets}
%     \link{Extended version}{https://aaai.org/example/extended-version}
% \end{links}

\section{Introduction}

The rapid advancement of generative models (GMs), especially GANs \cite{whichfaceisrealWhichFace} and DMs \cite{huang2023collaborative}, enables the creation of highly realistic synthetic images that are often indistinguishable from real photographs \cite{lu2023seeing}. While various detection methods have been proposed \cite{zhang2019detecting, liu2021spatial, frank2020leveraging, wang2023dire, tan2024rethinking}, their most critical limitation is the poor cross-generator generalization \cite{deng2024survey, yan2025a, yan2024orthogonal, chen2022ost}. \textbf{Cross-generator generalization}, in this context, refers to the ability of a detector trained on one set of GMs to successfully identify images produced by models it has never encountered. This failure is particularly pronounced when crossing the boundaries of underlying \textit{architectural families}, such as between GANs and DMs. This generalization gap is a critical obstacle, as new generators constantly emerge in the real world \cite{nguyen2025deepfake}.

To address this generalization gap, the fundamental challenge lies in understanding why different architectural families produce systematically different patterns. Prior research has primarily explored that GANs and DMs leave surface-level artifacts in synthetic images, such as checkerboard patterns \cite{dzanic2020fourier, zhang2019detecting, frank2020leveraging, Wang_2020_CVPR} and noise residuals \cite{wang2023dire, maexposing}. However, this approach faces an inherent limitation: these surface-level artifacts are vulnerable to simple post-processing operations, further limiting detection robustness. In contrast to prior work, we shift from analyzing surface-level artifacts to \textbf{latent architectural patterns} inherent to each generative architecture family. Specifically, we hypothesize that latent representations of synthetic images generated by GANs and DMs form significantly different submanifolds in feature space. This hypothesis stems from the fundamental observation that GANs and DMs have distinct optimization objectives, which fundamentally shape how these generative architecture families model data distribution: the Jensen-Shannon (JS) divergence is optimized by GANs, whereas the Kullback-Leibler (KL) divergence is minimized by DMs.

\begin{figure*}[ht]
    \centering
    \includegraphics[width=0.95\textwidth]{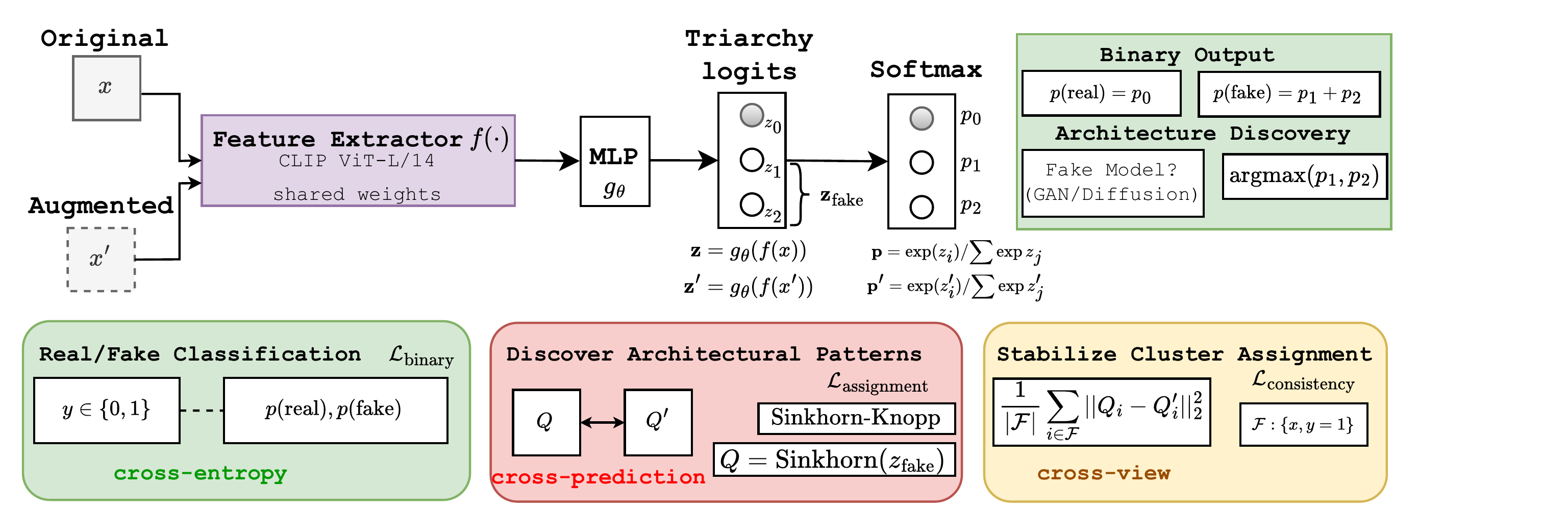}
    \caption{Overview of our proposed method, TriDetect.}
    \label{fig:overview}
\end{figure*}

\begin{figure*}[ht]
    \centering
    \includegraphics[width=0.95\textwidth]{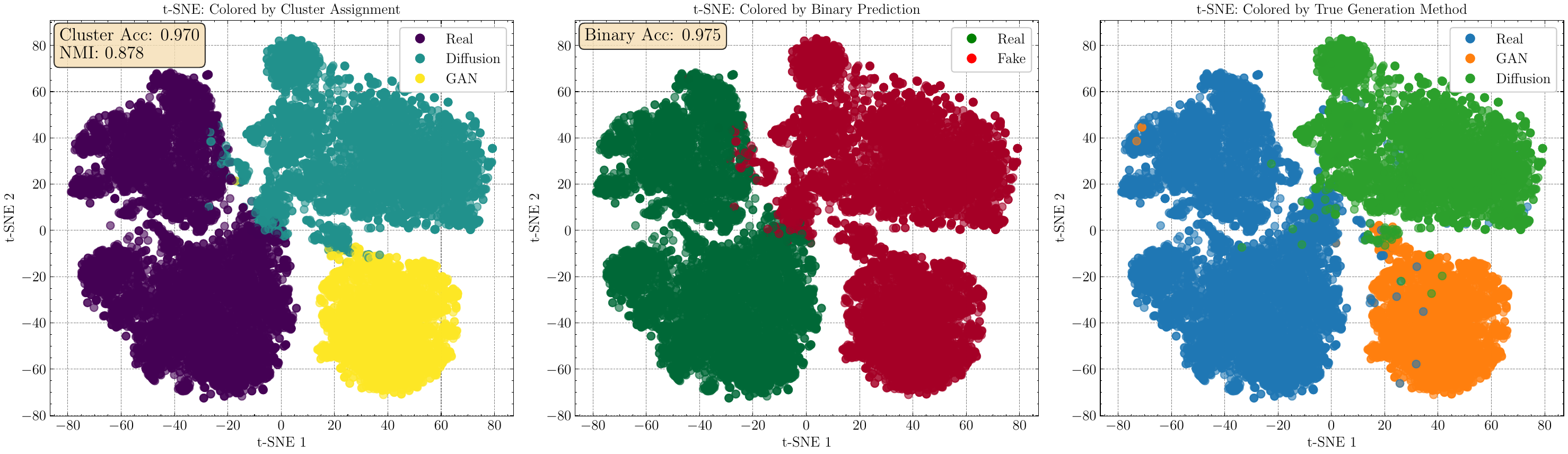}
    \caption{Visualization of learned representations demonstrating successful discovery of fake sub-types. The three t-SNE projections display feature embeddings colored by (left) the model's unsupervised cluster assignments, (middle) the model's binary real/fake predictions, and (right) the ground-truth generation methods. Results are performed on AIGCDetectBenchmark.}
    \label{fig:cluster-embed}
\end{figure*}

In this work, we demonstrate the existence of latent architectural patterns inherent to each generative architecture family (i.e., GANs and DMs) through a theoretical analysis of their optimization objectives. By utilizing the manifold hypothesis \cite{fefferman2016testing, loaizadeep}, we show that GANs and DMs interact with the data manifold differently: GANs permit partial manifold coverage, leading to boundary artifacts, while DMs enforce complete coverage, resulting in over-smoothing patterns. This theoretical disparity implies that GAN-generated and DM-generated images occupy distinct, separable submanifolds within a well-learned feature space (Fig. \ref{fig:cluster-embed}). This theoretical analysis directly guides our method design. Rather than attempting to detect specific generators or relying on surface-level patterns, we propose \textbf{Tri}archy \textbf{Detect}or (TriDetect), a novel semi-supervised method designed to recognize these fundamental architectural signatures. Specifically, TriDect simultaneously performs binary classification while discovering \textbf{latent architectural patterns} within the synthetic data (Fig. \ref{fig:overview}). To ensure robust learning, we employ the Sinkhorn-Knopp algorithm \cite{cuturi2013sinkhorn} to enforce balanced cluster assignments, preventing cluster collapse. Furthermore, a cross-view consistency mechanism is designed to ensure the model learns fundamental architectural distinctions rather than image statistics. By discovering latent architectural patterns, our method can enhance the cross-generator generalization within the same architecture families.

% Specifically, this clustering loss encourages the model to discover latent clusters corresponding to different generative architectures without requiring explicit architecture labels during training. It is important to note that this loss is specifically designed to enrich the learned representations rather than to perform architecture attribution. By compelling the model to learn features that can implicitly separate these architectural clusters, we facilitate the learning of more discriminative and generalized representations. This process improves the model's fundamental understanding of synthetic artifacts, thereby enhancing its ability to generalize across diverse and unseen generators within the same architectural family.

\textbf{Main Contributions.} Our contributions include:
\begin{itemize}
    \item \textbf{Theoretical Analysis}: We provide the first theoretical explanation for why the GAN and DM architectures produce fundamentally different latent patterns, grounded in their optimization objectives and manifold coverage.
    \item \textbf{Novel Semi-supervised Detection Method}: We introduce TriDetect that combines binary classification with architecture-aware clustering to learn generalized architectural features.
    \item \textbf{Comprehensive Evaluation}: We evaluate on 5 datasets against 13 SoTA detectors, demonstrating superior cross-generator and cross-dataset generalization.
\end{itemize}

\section{From Divergence to Detection: A Theoretical Analysis of Architectural Artifacts}
% Prior works have explored GAN and DM artifacts left by the generation process of GANs and DMs by empirically analyzing the upsampling module of GANs in the frequency domain \cite{frank2020leveraging,zhang2019detecting} and the denoising process of DMs \cite{wang2023dire}.

In this section, we provide the theoretical analysis to demonstrate that latent representations of synthetic images generated by GANs and DMs form significantly different submanifolds in feature space. The t-SNE visualization in Figure \ref{fig:cluster-embed} confirms this analysis. Specifically, our analysis reveals that these differences are fundamental consequences of their optimization targets. We begin by examining the optimization objectives of each architecture family, then demonstrate how these objectives lead to characteristic coverage patterns and to distinguishable artifacts.

\subsection{The Fundamental Disparity in Optimization}
We use $p_{data}$, $p_{GAN}$, and $p_{DM}$ to denote the data, GAN-generated, and DM-generated distributions, respectively.

\begin{lemma}(DM Optimization Objective).
    Suppose that $x_0 \sim p_{\text{data}}$, DM minimizes an upper bound on the KL divergence between the data and model distributions:
    \begin{equation}
        D_{\text{KL}}(p_{\text{data}} \| p_{\text{DM}}) \le \mathbb{E}_{x_0 \sim p_{\text{data}}}[-\text{ELBO}(x_0)] - H(p_{\text{data}}),
    \end{equation}
    where ELBO is the Evidence Lower Bound and $H(p_{\text{data}})$ is the entropy of the data distribution.
    \label{lemma:DM}
\end{lemma}

\begin{remark}(Equality Condition). The inequality in Lemma \ref{lemma:DM} becomes an equality when the variational posterior $q(x_{1:T}|x_0)$ equals the true posterior $p_{\text{DM}}(x_{1:T}|x_0)$ under the model. In standard DMs, this condition can be achieved at the global optimum when the reverse process has sufficient capacity \cite{luo2022understanding}, yielding:
\begin{equation}
    D_{\text{KL}}(p_{\text{data}} \| p_{\text{DM}}) = \mathbb{E}_{x_0 \sim p_{\text{data}}}[-\text{ELBO}(x_0)] + H(p_{\text{data}})
\end{equation}
 \label{remark:DM}
\end{remark}

 Since $H(p_{\text{data}})$ is independent of model parameters, minimizing $\mathbb{E}_{x_0 \sim p_{\text{data}}}[-\text{ELBO}(x_0)]$ is equivalent to minimizing $D_{KL}(p_{\text{data}} || p_{\text{DM}})$. Lemma \ref{lemma:DM} is supported by recent works on DMs \cite{luo2022understanding, nichol2021improved}, which indicates that DMs indirectly minimize the KL divergence by maximizing the ELBO. KL divergence is asymmetric and severely penalizes the model if $p_{DM}=0$ where $p_{data}>0$.

\begin{lemma}(GAN optimization problem)
    For a fixed generator $\mathcal{G}$, the value function $V(\mathcal{D}, \mathcal{G})$ satisfies the inequality:
    \begin{equation}
       V(\mathcal{D},\mathcal{G}) \leq  2D_{\text{JS}}(p_{\text{data}} \| p_{\text{GAN}}) - 2\log2,
    \end{equation}
    \label{lemma:GAN}
    with equality if and only if the discriminator is optimal, given by $\mathcal{D}^*(x) = p_{\text{data}}(x)/(p_{\text{data}}(x) + p_{\text{GAN}}(x))$
\end{lemma}

In contrast to DMs, GANs optimize the JS divergence through an adversarial game \cite{che2020your, goodfellow2020generative, arjovsky2017wasserstein}. Unlike the KL divergence, the JS divergence is symmetric and remains bounded even when the generator completely ignores certain regions of the data space.

These lemmas lead to our first key theoretical result:
\begin{theorem}(Fundamental Difference in Optimization Objectives)
    Using the notations of Lemmas \ref{lemma:DM} and \ref{lemma:GAN}, GANs and DMs solve various optimization problems:
    \begin{enumerate}
        \item GANs: $\min_{\mathcal{G}} V(\mathcal{D}^*,\mathcal{G}) \equiv  \min D_{\text{JS}}(p_{\text{data}} \| p_{\text{GAN}}), $ where $\mathcal{D}^{*}$ is the optimal discriminator.
        \item DMs: $\min \mathbb{E}_{x \sim p_{\text{data}}}[-\text{ELBO}(x)] -H(p_{\text{data}}) \equiv \min D_{\text{KL}}(p_{\text{data}} \| p_{\text{DM}})$
    \end{enumerate}
    \label{theorem:difference-optimization}
\end{theorem}

Proof for Theorem \ref{theorem:difference-optimization} in Appendix. Theorem \ref{theorem:difference-optimization} indicates the fundamental difference between the optimization objectives of GANs and DMs.

% remains bounded even when the generator completely ignores certain regions of the data space. This fundamental difference in penalty structure drives the distinct generative behaviors we observe.

\subsection{From Divergence to Distribution Support and Artifacts}
The choice of divergence profoundly impacts the learned distributions and the nature of artifacts inherent to the generation process.

\begin{theorem}(Disparity of Learned Distribution Support)
    Let $p_{\text{data}}$ be the true data distribution with support $S_{\text{data}} = \{x : p_{\text{data}}(x) > 0\}$. Assume that the generator and score-matching networks have sufficient capacity:
    \begin{enumerate}
        \item \textbf{GANs}: A globally optimal solution for the GAN generator, which minimizes the JS divergence $D_{\text{JS}}(p_{\text{data}} \| p_{\text{GAN}})$, can exist even if the support of the learned distribution, $S_{\text{GAN}}$,  is a strict subset of the data support ($S_{\text{GAN}} \subset S_{\text{data}}$, partial coverage).
        \item \textbf{DMs}: A globally optimal solution for the DM, which minimizes the upper bound on the KL divergence $D_{\text{KL}}(p_{\text{data}} \| p_{\text{DM}})$, requires the support of the learned distribution, $S_{\text{DM}}$, to cover the support of the data distribution ($S_{\text{data}} \subseteq S_{\text{DM}}$, complete coverage).
    \end{enumerate}
    \label{theorem:distribution-support}
\end{theorem}

Proof in Appendix. Theorem \ref{theorem:distribution-support} reveals why GANs and DMs exhibit fundamentally different coverage behaviors. For DMs, to achieve an optimal solution, the support of its learned distribution must cover the support of the real data distribution ($S_{\text{data}} \subseteq S_{\text{DM}}$). This is because the KL divergence $D_{\text{KL}}(p_{\text{data}} \| p_{\text{DM}})$ becomes infinite whenever $p_{\text{DM}} = 0$ for any $x$ where $p_{\text{data}} > 0$. As a result, DMs must spread their capacity across the entire data manifold, resulting in diffusion blur, over-smoothing in low-density regions. Furthermore, the iterative denoising process of DMs relies on score matching. Imperfections in this estimation accumulate during the reverse diffusion steps, combined with the smoothing enforced by the KL objective, will leave unique, structured noise residuals that serve as a distinct architectural artifact \cite{wang2023dire, maexposing}.
% forcing DMs to allocate probability mass across the entire data manifold. The denoising process introduces characteristic smoothness artifacts, particularly visible as blur patterns in high-frequency details.
In contrast, due to JS divergence optimization, GANs can achieve optimality while ignoring low-probability regions of $p_{\text{data}}$. This leads to sharp samples but incomplete coverage where GANs concentrate their capacity on high-density modes, producing sharp samples but missing rare patterns. The boundary artifacts manifest as abrupt transitions at the edges of learned support, creating detectable discontinuities. These discontinuities manifest in the image space as characteristic boundary artifacts, such as structural inconsistencies or unnatural texture transitions \cite{zhang2019detecting}.

\section{Methodology}
\label{sec:method}
% Traditional AI-generated image detectors treat all synthetic images as a \textit{single category}, learning binary decision boundaries that often fail to generalize across different generative architectures.
This section presents a semi-supervised \textbf{Tri}rchy \textbf{Detect}or (TriDetect) that simultaneously performs binary classification and discovers latent structures within synthetic images. From our theoretical analysis, by learning to recognize distinct architectural patterns that persist across specific generators within each family (i.e., GANs vs. DMs), it is possible to achieve cross-generator generalization capability.

\subsection{Problem Formulation and Objectives}
% Given a dataset $\mathcal{D} = \{(x_i, y_i)\}^N_{i=1}$ where $x_i \in \mathbb{R}^{H \times W \times 3}$ represents an image and $y_i \in \{0,1\}$ denotes its binary label (0:real, 1:fake), we extract semantic features $z_i = f(x_i) \in \mathbb{R}^d$ using a pre-trained vision-language foundation model $f$. We formulate a joint optimization problem with two complementary objectives: (i) distinguish real from synthetic images, and (ii) automatically discover latent clusters within synthetic samples that correspond to different generative architectures. To achieve these objectives, we employ an architecture-aware classifier $g_{\theta}: \mathbb{R}^d \rightarrow \mathbb{R}^N$ that produces logits for one real class and two fake clusters, which aligns with the theoretical distinction between GAN and diffusion model architectures established in our analysis.

Given a dataset $\mathcal{D} = \{(x_i, y_i)\}^N_{i=1}$ where $x_i \in \mathbb{R}^{H \times W \times 3}$ represents an image and $y_i \in \{0,1\}$ denotes its binary label (0:real, 1:fake), we compute logits $z_i = g_{\theta}(f(x_i)) \in \mathbb{R}^3$ using a fine-tuned vision encoder $f$ and classifier $g_{\theta}$, where we have 3 output dimensions (one real class and two fake clusters). We formulate a joint optimization problem with two objectives: (i) distinguish real from synthetic images, and (ii) automatically discover latent clusters within synthetic samples that correspond to different generative architectures. The architecture produces three-way logits that align with the theoretical distinction between GAN and DM architectures established in our analysis.

\subsection{Balanced Clustering via Optimal Transport}
A critical challenge in unsupervised deep clustering is the tendency toward trivial solutions where all samples collapse into a single cluster \cite{zhou2024comprehensive}. This problem is particularly serious in our setting, where we seek to discover subtle architectural patterns within synthetic images generated by different generators. To address this, we formulate cluster assignment as an optimal transport problem that enforces balanced distribution across clusters.

For a batch $B$ of synthetic samples with logits $Z_{\text{fake}} \in \mathbb{R}^{B \times K}$ where $K=2$ is the number of fake clusters, we seek an assignment matrix $Q \in \mathbb{R}^{B \times K}$ that maximizes the similarity between samples and clusters while maintaining balanced assignments:
\begin{equation}
    \max_{Q \in \mathcal{Q}} \text{Tr}(Q^T Z_{\text{fake}}) + \epsilon H(Q),
\end{equation}
where $H(Q) = - \sum_{ij} Q_{ij} \log Q_{ij}$ is the entropy function, $\epsilon$ controls assignment smoothness, and $\mathcal{Q}$ is the transportation polytope \cite{asano2019self}:
\begin{equation}
    \mathcal{Q} = \left\{ Q \in \mathbb{R}_{+}^{B \times K} \mid Q \mathbf{1}_K = \mathbf{1}_B, \; Q^T \mathbf{1}_B = \frac{B}{K} \mathbf{1}_K \right\}
\end{equation}

These constraints ensure that each sample is fully assigned across clusters (rows sum to 1) and each cluster receives exactly $B/K$ samples (columns sum to $B/K$), preventing collapse.

\textbf{Sinkhorn-Knopp Algorithm for Online Clustering.} We utilize the Sinkhorn-Knopp algorithm \cite{cuturi2013sinkhorn} to solve the optimal transport problem. Starting from the initialization $Q^{(0)} = \text{exp}(Z_{\text{fake}}/\epsilon)$, for $t = 1, \dots, T$ iterations, we perform:
\begin{equation}
    Q^{(t)} = \mathcal{R}(Q^{(t-1)}); \quad  Q^{(t)} = \mathcal{C}(Q^{(t)}),
    \label{eq:sinkhorn}
\end{equation}
where $\mathcal{R}(Q)_{ij} = \frac{Q_{ij}}{\sum_k Q_{ik}}$ normalizes rows and $\mathcal{C}(Q)_{ij} = \frac{K \cdot Q_{ij}}{\sum_{k} Q_{kj}}$ normalizes columns with appropriate scaling to maintain the balance constraints. Note that Eq. \ref{eq:sinkhorn} is operated on mini-batches rather than the entire dataset, enabling online learning that scales to arbitrarily large datasets.

\begin{algorithm}[h]
\small
\caption{TriDetect Training Algorithm}
\label{alg:TriDetect_training}
\begin{algorithmic}[1]
\Require Dataset $\mathcal{D} = \{(x_i, y_i)\}_{i=1}^N$, feature encoder $f$, classifier $g_\theta$, learning rate $\eta$, hyperparameters $\beta, \omega_1, \omega_2$, $K=2$ (fake clusters)
\Ensure Trained encoder $f$ and classifier $g_\theta$
\State Initialize classifier parameters $\theta$
\For{$t = 1$ to $T$ epochs}
    \For{each batch $\mathcal{B} \subset \mathcal{D}$}
        \State $\mathcal{B}' \gets \text{Augment}(\mathcal{B})$ \Comment{Create augmented views}
        \State $Z,Z' \gets g_\theta(f(\mathcal{B})), g_\theta(f(\mathcal{B}'))$ \Comment{Compute triarchy logits}
        \State $\mathcal{F} \gets \{i : y_i = 1\}$ \Comment{Extract fake indices}
        \State $Z_{\text{fake}}, Z'_{\text{fake}} \gets Z[\mathcal{F}, 1:K], Z'[\mathcal{F}, 1:K]$ \Comment{Extract fake logits}
        \State $Q_{\text{fake}}, Q'_{\text{fake}} \gets \text{Sinkhorn}(Z_{\text{fake}}), \text{Sinkhorn}(Z'_{\text{fake}})$ \Comment{Balanced assignment}
        \State Compute $\mathcal{L}_{\text{binary}}$ (Eq.\ref{eq:binary}), $\mathcal{L}_{\text{assignment}}$ (Eq.\ref{eq:assignment}), $\mathcal{L}_{\text{consistency}}$ (Eq.\ref{eq:consistency})
        \State $\mathcal{L}_{\text{cluster}} \gets \omega_1 \mathcal{L}_{\text{assignment}} + \omega_2 \mathcal{L}_{\text{consistency}}$
        \State $\mathcal{L}_{\text{total}} \gets \beta \mathcal{L}_{\text{binary}} + (1-\beta) \mathcal{L}_{\text{cluster}}$
        \State $\theta \gets \theta - \eta \nabla_\theta \mathcal{L}_{\text{total}}$ \Comment{Update parameters}
    \EndFor
\EndFor
\State \Return $f, g_\theta$
\end{algorithmic}
\end{algorithm}

\begin{table*}[ht]
\centering
\fontsize{7pt}{7pt}\selectfont
\caption{Comparison on the GenImage \cite{zhu2023genimage} Dataset in terms of AUC Performance. The best result and the second-best result are marked in \textbf{bold} and \ul{underline}, respectively.}
\label{table:GenImage-AUC}
\begin{tblr}{
  width = \linewidth,
  colspec = {Q[150]Q[108]Q[102]Q[88]Q[88]Q[110]Q[88]Q[110]Q[88]},
  cells = {c},
  row{15} = {FrenchPass},
  vline{2-9} = {-}{},
  hline{1,16} = {-}{0.08em},
  hline{2,15} = {-}{},
  hline{2} = {2}{-}{},
}
\textbf{Method} & \textbf{BigGAN} & \textbf{SD v1.4} & \textbf{ADM}    & \textbf{GLIDE}  & \textbf{MidJourney} & \textbf{VQDM}   & \textbf{Wukong} & \textbf{Avg}    \\
CNNSpot         & \textbf{1.0000} & \ul{0.9999}   & 0.8139          & 0.9591          & 0.7969              & 0.8028          & 0.9988          & 0.9102          \\
FreDect         & \textbf{1.0000} & 0.9970           & 0.7485          & 0.7713          & 0.7458              & 0.8086          & 0.9894          & 0.8658          \\
Fusing          & \ul{0.9999}  & \ul{0.9999}   & 0.7261          & 0.5664          & 0.6671              & 0.7586          & 0.9988          & 0.8167          \\
LGrad           & 0.6530          & 0.5623           & 0.5237          & 0.6103          & 0.5411              & 0.5244          & 0.6043          & 0.5742          \\
LNP             & 0.9812          & 0.9809           & 0.6090          & 0.6089          & 0.6596              & 0.6825          & 0.9689          & 0.7844          \\
CORE            & \textbf{1.0000} & 0.9998           & 0.8198          & 0.9950          & 0.8011              & 0.8192          & 0.9971          & 0.9189          \\
SPSL            & 0.9998          & 0.9970           & 0.5946          & 0.8283          & 0.7545              & 0.7742          & 0.9918          & 0.8486          \\
UIA-ViT         & \textbf{1.0000} & 0.9993           & 0.7314          & 0.9519          & 0.8024              & 0.7950          & 0.9946          & 0.8964          \\
DIRE            & 0.9996          & 0.9997           & 0.8257          & 0.9843          & 0.8298              & 0.8303          & 0.9965          & 0.9237          \\
UnivFD          & 0.9988          & 0.9955           & \ul{0.9265}  & 0.9920          & 0.9262              & 0.9859          & 0.9678          & 0.9704          \\
AIDE            & 0.9998          & 0.9900           & 0.7736          & 0.9299          & 0.8680              & 0.8739          & 0.9710          & 0.9152          \\
NPR             & \ul{0.9999}  & 0.9979           & 0.9385          & \ul{0.9956}  & 0.9430              & 0.9553          & 0.9861          & 0.9695          \\
Effort          & \textbf{1.0000} & \textbf{1.0000}  & 0.9052          & 0.9893          & \textbf{0.9795}     & \ul{0.9969}  & \ul{0.9998}  & \ul{0.9815}  \\
TriDetect             & \textbf{1.0000} & \textbf{1.0000}  & \textbf{0.9609} & \textbf{0.9993} & \ul{0.9581}      & \textbf{0.9992} & \textbf{0.9999} & \textbf{0.9882}
\end{tblr}
\end{table*}

\begin{table*}[ht]
\centering
\fontsize{6pt}{6pt}\selectfont
\caption{Comparison on the AIGCDetectBenchmark \cite{zhong2023patchcraft} Dataset in terms of AUC Performance. The best result and the second-best result are marked in \textbf{bold} and \ul{underline}, respectively.}
\label{table:AIGC-AUC}
\begin{tblr}{
  width = \linewidth,
  colspec = {Q[l, 80] *{17}{Q[c, 52]}}, % Centered columns
  cells = {c},
  row{1} = {cmd=\rotatebox{65}}, % Apply rotation to all cells in the first row
  row{15} = {FrenchPass},
  vline{2-18} = {-}{},
  hline{1,16} = {-}{0.08em},
  hline{2,15} = {-}{},
  hline{2} = {2}{-}{},
}
\textbf{Method} & \textbf{CycleGAN} & \textbf{StyleGAN} & \textbf{StyleGAN2} & \textbf{ProGAN} & \textbf{GauGAN} & \textbf{StarGAN} & \textbf{BigGAN} & \textbf{ADM}    & \textbf{Wukong} & \textbf{GLide}  & \textbf{SD\_XL} & \textbf{VQDM}   & \textbf{MidJourney} & \textbf{SD v1.4} & \textbf{SD v1.5} & \textbf{DALLE2} & \textbf{Avg}    \\
CNNSpot         & 0.3491            & 0.5298            & 0.4787             & 0.5039          & 0.4260          & 0.5538           & 0.3811          & 0.8138          & 0.9988          & 0.9591          & 0.8252          & 0.8028          & 0.7969              & 0.9999           & 0.9995           & 0.9748          & 0.7121          \\
FreDect         & 0.3956            & 0.6697            & 0.6815             & 0.7172          & 0.5877          & 0.7211           & 0.8346          & 0.7485          & 0.9894          & 0.7713          & 0.8743          & 0.8086          & 0.7458              & 0.9970           & 0.9956           & 0.8224          & 0.7725          \\
Fusing          & 0.4071            & 0.5180            & 0.5274             & 0.4797          & 0.4815          & 0.5072           & 0.3888          & 0.7261          & 0.9988          & 0.9429          & 0.7468          & 0.7586          & 0.6673              & \ul{0.9999}   & 0.9998           & 0.9448          & 0.6934          \\
LGrad           & 0.4833            & 0.4783            & 0.4181             & 0.4993          & 0.4508          & 0.4925           & 0.4972          & 0.5317          & 0.6009          & 0.6758          & 0.4593          & 0.5100          & 0.5320              & 0.5639           & 0.5617           & 0.6041          & 0.5224          \\
LNP             & 0.3984            & 0.5118            & 0.4842             & 0.4993          & 0.4884          & 0.4283           & 0.4246          & 0.6089          & 0.9689          & 0.8474          & 0.7127          & 0.6823          & 0.6596              & 0.9809           & 0.9800           & 0.8382          & 0.6571          \\
CORE            & 0.6153            & 0.6115            & 0.5762             & 0.5617          & 0.5310          & 0.9731           & 0.5574          & 0.8198          & \ul{0.9971}  & \textbf{0.9950} & 0.7561          & 0.8193          & 0.8011              & 0.9998           & 0.9997           & 0.9498          & 0.7852          \\
SPSL            & 0.5817            & 0.5384            & 0.5305             & 0.5252          & 0.5209          & 0.5228           & 0.6458          & 0.5946          & 0.9918          & 0.8283          & 0.8257          & 0.7742          & 0.7545              & 0.9971           & 0.9953           & 0.8301          & 0.7161          \\
UIA-ViT         & 0.5291            & 0.7081            & 0.7545             & 0.5109          & 0.4934          & 0.5252           & 0.6471          & 0.7314          & 0.9946          & 0.9519          & 0.9358          & 0.7950          & 0.8024              & 0.9993           & 0.9988           & 0.8187          & 0.7623          \\
DIRE            & 0.6335            & 0.5967            & 0.6117             & 0.7896          & 0.4016          & 0.9935           & 0.5755          & 0.8257          & 0.9965          & 0.9843          & 0.8928          & 0.8303          & 0.8298              & 0.9997           & \ul{0.9995}   & 0.9627          & 0.8077          \\
UnivFD          & 0.9664            & 0.7297            & 0.7304             & 0.9407          & 0.9828          & 0.9185           & 0.9712          & 0.9265          & 0.9678          & \ul{0.9920}  & 0.9610          & 0.9859          & 0.9262              & 0.9955           & 0.9953           & 0.9627          & 0.9345          \\
AIDE            & 0.6142            & 0.7377            & 0.7249             & 0.7979          & 0.8443          & 0.7025           & 0.8084          & 0.7736          & 0.9710          & 0.9299          & 0.9309          & 0.8739          & 0.8680              & 0.9900           & 0.9900           & 0.8859          & 0.8402          \\
NPR             & 0.9288            & 0.8851            & \ul{0.9403}     & 0.9653          & 0.5391          & \textbf{0.9998}  & 0.7939          & \textbf{0.9385} & 0.9861          & 0.9956          & 0.9728          & 0.9553          & 0.9430              & 0.9979           & 0.9967           & \ul{0.9935}  & 0.9270          \\
Effort          & \ul{0.9894}    & \textbf{0.9431}   & 0.8961             & \ul{0.9973}  & \textbf{0.9997} & 0.9923           & \ul{0.9991}  & 0.9052          & \ul{0.9998}  & 0.9893          & \textbf{0.9980} & \ul{0.9969}  & \textbf{0.9795}     & \textbf{1.0000}  & \textbf{1.0000}  & 0.9666          & \ul{0.9783}  \\
TriDetect             & \textbf{1.0000}   & \ul{0.9419}    & \textbf{0.9422}    & \textbf{0.9995} & \ul{0.9978}  & \textbf{1.0000}  & \textbf{0.9992} & \textbf{0.9609} & \textbf{0.9999} & \textbf{0.9993} & \ul{0.9948}  & \textbf{0.9992} & \ul{0.9581}      & \textbf{1.0000}  & \textbf{1.0000}  & \textbf{0.9971} & \textbf{0.9869}
\end{tblr}
\end{table*}

\begin{table*}[ht]
\centering
\fontsize{7pt}{7pt}\selectfont
\caption{Comparison on the WildFake \cite{hong2025wildfake} Dataset in terms of ACC Performance. The best result and the second-best result are marked in \textbf{bold} and \ul{underline}, respectively.}
\label{table:wildfake-ACC}
\begin{tblr}{
  width = \linewidth,
  colspec = {Q[87]Q[77]Q[67]Q[67]Q[67]Q[81]Q[85]Q[92]Q[81]Q[67]Q[88]Q[67]},
  cells = {c},
  row{15} = {FrenchPass},
  vline{2-12} = {-}{},
  hline{1,16} = {-}{0.08em},
  hline{2,15} = {-}{},
  hline{2} = {2}{-}{},
}
\textbf{Method} & \textbf{DALL-E} & \textbf{DDIM}   & \textbf{DDPM}   & \textbf{VQDM}   & \textbf{BigGAN} & \textbf{StarGAN} & \textbf{StyleGAN} & \textbf{DF-GAN} & \textbf{GALIP}  & \textbf{GigaGAN} & \textbf{Avg}    \\
CNNSpot         & 0.5961          & 0.2929          & 0.2811          & 0.2882          & 0.8456          & 0.3350           & 0.3453            & 0.3418          & 0.3325          & 0.3409           & 0.3999          \\
FreDect         & 0.4078          & 0.2817          & 0.2952          & 0.4643          & 0.8415          & 0.3689           & 0.3365            & 0.4415          & 0.3478          & \ul{0.8148}   & 0.4600          \\
Fusing          & 0.4706          & 0.2376          & 0.2427          & 0.3305          & \textbf{0.9361} & 0.3425           & 0.3361            & 0.3518          & 0.3587          & 0.3446           & 0.3951          \\
LGrad           & \ul{0.6044}  & 0.5794          & 0.5473          & 0.5656          & 0.6126          & 0.5517           & 0.5129            & 0.5732          & 0.4796          & 0.5758           & 0.5602          \\
LNP             & 0.5105          & 0.2681          & 0.2715          & 0.3580          & 0.7179          & 0.3741           & 0.3696            & 0.3697          & 0.3649          & 0.3708           & 0.3975          \\
CORE            & 0.3829          & 0.2552          & 0.2762          & 0.3668          & 0.7115          & 0.3359           & 0.3353            & 0.3974          & 0.3365          & 0.3400           & 0.3738          \\
SPSL            & 0.5614          & 0.2282          & 0.2535          & 0.3082          & 0.8692          & 0.3324           & 0.3395            & 0.3618          & 0.3338          & 0.3334           & 0.3921          \\
UIA-ViT         & 0.5718          & 0.2699          & 0.2636          & 0.3475          & 0.7542          & 0.3402           & 0.3635            & 0.3806          & 0.3430          & 0.3592           & 0.3993          \\
DIRE            & 0.5725          & 0.3222          & 0.3130          & 0.4370          & 0.6130          & 0.3885           & 0.3465            & 0.3941          & 0.3724          & 0.3755           & 0.4135          \\
UnivFD          & 0.5492          & \textbf{0.8244} & \textbf{0.7536} & 0.7974          & 0.7962          & 0.6213           & 0.5734            & 0.9141          & \ul{0.8572}  & 0.7690           & 0.7456          \\
AIDE            & 0.5806          & 0.3562          & 0.3405          & 0.4757          & 0.7420          & 0.3948           & 0.4331            & 0.6638          & 0.4215          & 0.3647           & 0.4773          \\
NPR             & 0.5122          & 0.6892          & 0.5305          & 0.7313          & 0.7102          & 0.5510           & 0.3583            & 0.7753          & 0.5221          & 0.6876           & 0.6068          \\
Effort          & \textbf{0.6741} & 0.6436          & 0.4800          & \ul{0.8467}  & \ul{0.8992}  & \ul{0.7971}   & \ul{0.6027}    & \ul{0.9767}  & 0.8127          & 0.7891           & \ul{0.7522}  \\
TriDetect             & 0.6021          & \ul{0.7359}  & \ul{0.7207}  & \textbf{0.9301} & 0.8042          & \textbf{0.9998}  & \textbf{0.6300}   & \textbf{0.9825} & \textbf{0.9017} & \textbf{0.9468}  & \textbf{0.8254}
\end{tblr}
\end{table*}

\begin{table*}[ht]
\centering
\fontsize{7pt}{7pt}\selectfont
\caption{Comparison on the DF40 \cite{yan2024df40} Dataset in terms of ACC Performance. The best result and the second-best result are marked in \textbf{bold} and \ul{underline}, respectively.}
\label{table:DF40-ACC}
\begin{tblr}{
  width = \linewidth,
  colspec = {Q[130]Q[102]Q[117]Q[100]Q[92]Q[102]Q[100]Q[100]Q[73]},
  cells = {c},
  row{15} = {FrenchPass},
  vline{2-9} = {-}{},
  hline{1,16} = {-}{0.08em},
  hline{2,15} = {-}{},
  hline{2} = {2}{-}{},
}
\textbf{Method} & \textbf{CollabDiff} & \textbf{MidJourney} & \textbf{DeepFaceLab} & \textbf{StarGAN} & \textbf{StarGAN2} & \textbf{StyleCLIP} & \textbf{WhichisReal} & \textbf{Avg}    \\
CNNSpot         & 0.5635              & 0.5376              & 0.3808               & 0.4851           & 0.5040            & 0.3041             & 0.5012               & 0.4681          \\
FreDect         & 0.8720              & 0.4614              & 0.3322               & 0.4982           & 0.5158            & 0.4103             & 0.5327               & 0.5175          \\
Fusing          & 0.6150              & 0.5296              & 0.4830               & 0.4554           & 0.4942            & 0.2908             & 0.4978               & 0.4808          \\
LGrad           & 0.5045              & 0.5019              & 0.5374               & 0.5088           & 0.4670            & 0.6124             & 0.5107               & 0.5204          \\
LNP             & 0.5120              & 0.4997              & 0.3808               & 0.4962           & 0.5108            & 0.3428             & 0.5292               & 0.4674          \\
CORE            & 0.5025              & 0.5328              & 0.5708               & 0.5000           & 0.5000            & 0.2612             & 0.4998               & 0.4810          \\
SPSL            & 0.5140              & 0.5280              & 0.3337               & 0.4955           & 0.4992            & 0.2884             & 0.4988               & 0.4511          \\
UIA-ViT         & 0.5380              & 0.4374              & 0.3609               & 0.5033           & 0.5008            & 0.3930             & 0.4698               & 0.4576          \\
DIRE            & 0.5070              & 0.5424              & 0.3275               & 0.5000           & 0.5000            & 0.2652             & 0.4658               & 0.4440          \\
UnivFD          & 0.5955              & 0.7304              & 0.5369               & 0.6361           & 0.5516            & 0.8353             & 0.6727               & 0.6512          \\
NPR             & 0.4995              & \textbf{0.8711}     & 0.3228               & 0.4970           & 0.5000            & 0.2610             & 0.6442               & 0.5136          \\
AIDE            & 0.5525              & 0.7038              & 0.4039               & 0.6331           & 0.5055            & 0.4066             & 0.5197               & 0.5322          \\
Effort          & \textbf{0.9735}     & 0.8370              & \ul{0.4500}       & \textbf{0.9287}  & \ul{0.7345}    & \ul{0.9710}     & \textbf{0.8291}      & \ul{0.8177}  \\
TriDetect         & \ul{0.9690}      & \ul{0.8629}      & \textbf{0.6402}      & \ul{0.8005}   & \textbf{0.8208}   & \textbf{0.9825}    & \ul{0.8241}       & \textbf{0.8429}
\end{tblr}
\end{table*}

\subsection{Cross-view Consistency for Generalized Learning}
To learn discriminative and generalized clusters that capture architectural patterns rather than image statistics, we propose a cross-view consistency mechanism that encourages consistent predictions and clustering stability. This mechanism contains the \textit{swapped prediction strategy} and \textit{consistency regularization}.

\textbf{Swapped Prediction for Cluster Learning.} Let $(x, x')$ denote different views, generated by random transformations on each image, with corresponding logits $(z, z')$. For fake samples, we extract the fake cluster logits $z_{\text{fake}}$ and $z'_{\text{fake}}$. Inspired by the work \cite{caron2020unsupervised}, the core idea here is to use the cluster assignment from one view to supervise the prediction from the other view. Specifically, after computing balanced assignments using the Sinkhorn-Knopp algorithm $Q = \text{Sinkhorn}(z_{\text{fake}})$ and $Q' = \text{Sinkhorn}(z'_{\text{fake}})$ for fake samples in each view, the assignment loss is computed as follows:
\begin{equation}
\resizebox{0.9\linewidth}{!}{
  $\displaystyle
  \mathcal{L}_{\text{assignment}} = - \frac{1}{2|\mathcal{F}|} \sum_{i \in \mathcal{F}} \left( \sum_{k=1}^{K} Q'_{ik} \log p_{ik} + \sum_{k=1}^{K} Q_{ik} \log p'_{ik} \right),
  $
}
\label{eq:assignment}
\end{equation}
where $\mathcal{F}$ denotes the set of fake samples in the batch and $p_{ik} = \frac{\exp(z_{ik}/\tau)}{\sum_j \exp(z_{ij}/\tau)}$ with temperature $\tau$.

\textbf{Consistency Regularization for Stable Assignments.} The consistency regularization ensures that the Sinkhorn-Knopp assignments themselves remain stable across views, preventing oscillation during training and encouraging convergence to meaningful clusters. Particularly, this additional regularization directly constrains the similarity of assignments themselves:
\begin{equation}
    \mathcal{L}_{\text{consistency}} = \frac{1}{|\mathcal{F}|} \sum_{i \in \mathcal{F}} \|Q_i - Q'_i\|_2^2
    \label{eq:consistency}
\end{equation}
%The distinction between these losses is crucial: the assignment loss teaches the model what to predict (learning discriminative features for clustering), while the consistency loss ensures how consistently to assign (stabilizing the clustering process itself).

\subsection{Loss Function}
The overall loss function for training TriDetect contains the binary classification loss and the clustering loss:
\begin{equation}
    \mathcal{L}_{\text{total}} = \beta \mathcal{L}_{\text{binary}} + (1-\beta) \mathcal{L}_{\text{cluster}},
\end{equation}
where $\mathcal{L}_{\text{cluster}} = \omega_1 \mathcal{L}_{\text{assignment}} + \omega_2 \mathcal{L}_{\text{consistency}}$.
% The clustering loss aggregates both assignment and consistency regularization components:
% \begin{equation}
%     \mathcal{L}_{\text{cluster}} = \omega_1 \mathcal{L}_{\text{assignment}} + \omega_2 \mathcal{L}_{\text{consistency}}
% \end{equation}
$\beta$ denotes the hyperparameter that balances the importance of the binary and clustering losses while $\omega_1$ and $\omega_2$ represent assignment and consistency weights. $\mathcal{L}_{\text{binary}}$ is the cross-entropy loss, computed as:
% \begin{equation}
%     \mathcal{L}_{\text{binary}} = -\frac{1}{N}\sum_{i=1}^N [y_i \log p(\text{fake})_i + (1-y_i)\log p(\text{real})_i],
%     \label{eq:binary}
% \end{equation}
% where $p(\text{fake}) = \text{softmax}(z_1) + \text{softmax}(z_2)$ and $p(\text{real}) = \text{softmax}(z_0)$.

\begin{equation}
    \mathcal{L}_{\text{binary}} = -\frac{1}{N}\sum_{i=1}^N [y_i \log p(\text{fake})_i + (1-y_i)\log p(\text{real})_i],
    \label{eq:binary}
\end{equation}
where $p_i = \text{softmax}(z_i) \in \mathbb{R}^3$, with $p(\text{real})_i = p_{i,0}$ and $p(\text{fake})_i = p_{i,1} + p_{i,2}$. We provide the training pseudocode in Alg. \ref{alg:TriDetect_training} and the Sinkhorn-Knopp algorithm in Appendix.

\begin{table}[ht]
\centering
\fontsize{7pt}{7pt}\selectfont
\caption{Comparison on the Chameleon \cite{yan2025a} Dataset in terms of AUC, ACC, EER, and AP Performance. The best result and the second-best result are marked in \textbf{bold} and \ul{underline}, respectively.}
\label{table:chameleon}
\begin{tblr}{
  width = \linewidth,
  colspec = {Q[225]Q[169]Q[169]Q[169]Q[169]},
  cells = {c},
  row{15} = {FrenchPass},
  vline{2} = {-}{},
  hline{1,16} = {-}{0.08em},
  hline{2,15} = {-}{},
  hline{2} = {2}{-}{},
}
\textbf{Method} & \textbf{AUC}    & \textbf{ACC}    & \textbf{EER}    & \textbf{AP}     \\
CNNSpot         & 0.6331          & 0.5838          & 0.4062          & 0.5800          \\
FreDect         & 0.7545          & 0.6304          & 0.3143          & 0.6864          \\
Fusing          & 0.6020          & 0.5745          & 0.4138          & 0.5350          \\
LGrad           & 0.5342          & 0.5155          & 0.4809          & 0.4735          \\
LNP             & 0.5798          & 0.5790          & 0.4420          & 0.5039          \\
CORE            & 0.5264          & 0.5700          & 0.4772          & 0.4556          \\
SPSL            & 0.6135          & 0.5748          & 0.4147          & 0.5335          \\
UIA-ViT         & 0.7042          & 0.5987          & 0.3538          & 0.6115          \\
DIRE            & 0.5488          & 0.5768          & 0.4565          & 0.4909          \\
UnivFD          & 0.8030          & \ul{0.6598} & 0.2670          & 0.7506          \\
AIDE            & 0.7970          & 0.6256          & 0.2747          & 0.7345          \\
NPR             & 0.5865          & 0.5805          & 0.4381          & 0.4943          \\
Effort          & \ul{0.8371}  & 0.6459          & \ul{0.2428}  & \ul{0.7692}  \\
TriDetect             & \textbf{0.8935} & \textbf{0.6788}  & \textbf{0.1843} & \textbf{0.8742}
\end{tblr}
\end{table}

\section{Experiments}
\subsection{Experimental Setup}
\textbf{Baselines.} We compare our method against 13 competitive detectors, including CNNSpot \cite{wang2020cnn}, LNP \cite{liu2022detecting}, FreDect \cite{frank2020leveraging}, CORE \cite{ni2022core}, SPSL \cite{liu2021spatial}, UIA-ViT \cite{zhuang2022uia}, Fusing \cite{ju2022fusing}, LGrad \cite{tan2023learning}, DIRE \cite{wang2023dire}, UnivFD \cite{ojha2023towards}, AIDE \cite{yan2025a}, NPR \cite{tan2024rethinking}, and Effort \cite{yan2024orthogonal}.

\textbf{Datasets.} We evaluate our method on 2 standard benchmarks (GenImage \cite{zhu2023genimage}, AIGCDetectBenchmark \cite{zhong2023patchcraft}) and 3 in-the-wild datasets (WildFake \cite{hong2025wildfake}, Chameleon \cite{yan2025a}, DF40 \cite{yan2024df40}). In our experiments, we follow the training protocol introduced by \cite{yan2025a} by utilizing BigGAN and SDv1.4 subsets of GenImage.

% we utilize the training protocol introduced by \cite{yan2025a}, which demonstrated that training on a mixture of GAN and DM data significantly improves generalization capability compared to training on a single architecture. Following this work, we use BigGAN and SDv1.4 subsets of GenImage as the training dataset for all methods.

% This generalization failure stems from a fundamental limitation: current approaches typically train models exclusively on images generated by Generative Adversarial Networks (GANs) and evaluate their performance on datasets containing images from various generative architectures, including both GANs and Diffusion Models (DMs). As revealed by \citeauthor{yan2025a} \cite{yan2025a}, this training-evaluation mismatch imposes an impractical constraint that fundamentally hinders the model's ability to learn from the diverse properties exhibited by different generative architectures.

\textbf{Metrics.} In this work, we report results across 4 commonly used metrics: accuracy (ACC), area under the ROC curve (AUC), equal error rate (EER), and average precision (AP). We primarily report AUC and ACC in the main paper; other metric results are provided in Appendix.
% All results are averaged over both real and AI-generated images unless otherwise specified.

\textbf{Implementation Details.} We use CLIP ViT-L/14 \cite{radford2021learning} vision encoder $f$, which is fine-tuned with Low-Rank Adaptation (LoRA) \cite{hulora}. For baseline implementation, we use two benchmarks, DeepfakeBench \cite{yan2023deepfakebench} and AIGCDetectBenchmark \cite{zhong2023patchcraft}. Detailed experimental procedures, hyperparameters, and configurations are provided in Appendix.

% This choice serves dual purposes: (i) preserving the model's general knowledge about real images while avoiding catastrophic forgetting of pre-trained representations \cite{biderman2024lora}; and (ii) reducing the number of trainable parameters, making our approach \textcolor{red}{computationally efficient and scalable}.

\subsection{Experimental Results}
\subsubsection{Standard Benchmarks.}
% Tables \ref{table:GenImage-AUC} and \ref{table:AIGC-AUC} show the quantitative results of our method on two standard benchmarks.
On GenImage, Table \ref{table:GenImage-AUC} shows that TriDetect achieves the highest average AUC of 0.9882, outperforming SoTA methods, including AIDE (0.9152), NPR (0.9695), and Effort (0.9815). The results on AIGCDetectBenchmark (Table \ref{table:AIGC-AUC}) also demonstrate the generalization capability of TriDetect to 16 unseen generators, yielding an improvement over DIRE (22.18\%), AIDE (17.4\%), UnivFD (5.6\%), NPR (6.5\%), and Effort (0.9\%).

\subsubsection{In-the-wild Datasets.}
Tables \ref{table:wildfake-ACC}-\ref{table:chameleon} show that our method outperforms existing detectors across all evaluation metrics on in-the-wild datasets:
\begin{itemize}
    \item \textbf{Robust to degradation techniques:} TriDetect exhibits consistent performance when evaluated on WildFake in which degradation techniques (e.g., downsampling, cropping) are applied on testing sets. Particularly, TriDetect achieves an average ACC of 0.8254, a $8.86\%$ improvement compared to Effort. Results demonstrate the better robustness of TriDetect than other methods.
    \item \textbf{Generalize to facial editing methods:} On DF40 dataset, TriDetect achieves an average ACC of 0.8429, an improvement of 3.08\% and 29.43\% compared to Effort and UnivFD, respectively.
    \item \textbf{Achieve lowest EER:} On the challenging Chameleon dataset (Table \ref{table:chameleon}), which is designed to deceive both humans and AI models, TriDetect attains the lowest EER of 0.1843, a 31.74\% reduction compared to Effort.  This demonstrates that our approach can achieve a superior balance between false positive and false negative rates.
\end{itemize}

\begin{table}[ht]
\centering
\fontsize{7pt}{7pt}\selectfont
\caption{Ablation study on the binary classification loss ($\mathcal{L}_{\text{binary}}$), assignment loss ($\mathcal{L}_{\text{assignment}}$), and consistency loss ($\mathcal{L}_{\text{consistency}}$). Results are performed on Chameleon dataset.}
\label{ablation:module}
\begin{tblr}{
  width = \linewidth,
  colspec = {Q[100]Q[120]Q[120]Q[62]Q[62]Q[62]Q[62]},
  cells = {c},
  vline{2-7} = {-}{},
  hline{1,5} = {-}{0.08em},
  hline{2} = {-}{},
  hline{2} = {2}{-}{},
}
\textbf{$\mathcal{L}_{\text{binary}}$} & \textbf{$\mathcal{L}_{\text{assignment}}$} & \textbf{$\mathcal{L}_{\text{consistency}}$} & \textbf{AUC}    & \textbf{ACC}    & \textbf{EER}    & \textbf{AP}     \\
\checkmark                                  &                                                 &                                              & 0.8033          & 0.6335          & 0.2713          & 0.7503          \\
\checkmark                                  & \checkmark                                      &                                              & 0.8560          & 0.5817          & 0.2195          & 0.8394          \\
\checkmark                                  & \checkmark                                      & \checkmark                                   & \textbf{0.8935} & \textbf{0.6258} & \textbf{0.1843} & \textbf{0.8742}
\end{tblr}
\end{table}

\begin{table}[ht]
\centering
\fontsize{7pt}{7pt}\selectfont
\caption{Ablation studies on $\beta$ value in our method. Results are performed on Chameleon dataset.}
\label{ablation:beta}
\begin{tblr}{
  width = \linewidth,
  colspec = {Q[348]Q[138]Q[140]Q[140]Q[140]},
  cells = {c},
  vline{2} = {-}{},
  hline{1,6} = {-}{0.08em},
  hline{2} = {-}{},
  hline{2} = {2}{-}{},
}
\textbf{$\beta$ value} & \textbf{AUC}    & \textbf{ACC}    & \textbf{EER}    & \textbf{AP}     \\
$\beta=0.3$            & 0.8423          & 0.6069          & 0.2348          & 0.8147          \\
$\beta=0.5$            & 0.8451          & 0.5994          & 0.2338          & 0.8058          \\
$\beta=0.7$            & \textbf{0.8935} & \textbf{0.6258} & \textbf{0.1843} & \textbf{0.8742} \\
$\beta=0.9$            & 0.8774          & 0.6043          & 0.2033          & 0.8713
\end{tblr}
\end{table}

\section{Ablation Study}
\subsection{Module Ablation of Components in TriDetect}

Table \ref{ablation:module} provides evidence for the contribution of each loss component in our TriDetect. With $\mathcal{L}_{\text{assignment}}$ and $\mathcal{L}_{\text{consistency}}$, TriDetect can yield a substantial improvement in AUC to 0.8935 and ACC to 0.6258. These results highlight the significant contribution of the clustering loss to TriDetect's performance.
% Furthermore, this validates the core hypothesis that discovering latent structure within fake samples enhances generalization capabilities.

\subsection{Different Values of $\beta$}
The ablation study on $\beta$ (Table \ref{ablation:beta}) reveals a crucial insight: the optimal balance between binary classification and clustering objectives occurs at $\beta=0.7$, where the model achieves peak performance across all metrics.

\subsection{Analysis on Learned Embedding Spaces}
Figure \ref{fig:cluster-embed} provides visual evidence that TriDetect successfully discovers meaningful latent structure within synthetic images. The alignment between unsupervised discovery (left figure) and ground truth (right figure) validates our theoretical analysis that different generative architectures leave distinguishable artifacts in feature space. The middle figure further confirms that this clustering capability enhances rather than compromises binary detection performance.

We also provide other ablation studies in Appendix, including comparison with attribution baselines and ablation study on different numbers of fake clusters. Results show that TriDetect outperforms the selected attribution baselines, and $K=2$ is the optimal number of fake clusters.

\section{Related Work}
\label{sec:related-work}
% In general, there are two types of synthetic images: (i) \textbf{entirely AI-generated} images created without human subjects as input, and (ii) \textbf{deepfake} images designed for impersonation purposes. This section provides an overview of recent advances in generalizable detection methods for both categories.

\subsection{Generalized AI-generated Image Detection}
AI-generated images exhibit artifacts throughout the entire scene. The detection landscape encompasses three primary approaches: data augmentations for improved generalization \cite{wang2020cnn}, artifact analysis techniques identifying generation-specific patterns such as GAN checkerboard artifacts \cite{zhang2019detecting, tan2024rethinking} and diffusion reconstruction errors \cite{luo2024lare, wang2023dire}, and vision-language models leveraging CLIP representations to capture subtle synthetic patterns \cite{ojha2023towards, yansanity, yan2024orthogonal}. Recent advances include the exploitation of upsampling traces common to both GANs and DMs \cite{tan2024rethinking}, while adaptation strategies employ meta-learning \cite{chen2022ost}, and incremental learning \cite{pan2023dfil} to address evolving generators.

Compared to these works, we provide the first theoretical explanation for distinct artifacts left by GANs and DMs. These theoretical results motivate us to design TriDetect that simultaneously performs binary detection and discovers architectural patterns without explicit supervision.

\subsection{Deep Clustering}
Deep clustering has evolved from multi-stage pipelines separating representation learning and clustering \cite{tao2021clustering, zhang2021learning, peng2016deep} to end-to-end approaches jointly optimizing both objectives \cite{ji2021variational, shen2021you, caron2020unsupervised, qian2023stable}. Notable advances include utilizing the Sinkhorn-Knopp algorithm for optimal cluster assignment through prototype matching \cite{caron2020unsupervised} and designing hardness-aware objectives for training stability \cite{qian2023stable}.

While our method also employs Sinkhorn-Knopp, we uniquely combine it with semi-supervised learning to leverage real image supervision while discovering architectural patterns in synthetic images. We also propose a consistency regularization loss to ensure stable cluster assignment.

% Deep clustering has emerged as a powerful paradigm for unsupervised learning. Early works often employed a multi-stage pipeline, first learning a deep representation before applying a separate clustering algorithm \cite{tao2021clustering, zhang2021learning, peng2016deep}. Recently, end-to-end clustering approaches \cite{ji2021variational, shen2021you, caron2020unsupervised} have become dominant, which jointly optimize the representation learning module and the clustering module. SwAV \cite{caron2020unsupervised} utilizes Sinkhorn-Knopp algorithm to compute the cluster assignments by optimally matching the batch's feature vectors to a set of trainable prototypes. SeCu \cite{qian2023stable} introduces a hardness-aware learning objective that stabilizes training by modifying the cross-entropy loss to stop the gradient from negative instances when updating cluster centers.

\section{Conclusion}
% In this work, we provide the first theoretical analysis explaining why GANs and DMs produce different artifacts, which can be utilized for AI-generated image detection. This insight motivates TriDetect, a semi-supervised detection method that enhances binary classification with architecture-aware clustering. Through discovering latent architectural patterns without explicit labels, TriDetect can learn more generalized representations that help to improve the generalization capabilities across unseen generators within the same architecture families. This work establishes a new paradigm for AI-generated image detection, shifting from pattern recognition to architectural understanding, providing a foundation for addressing future generative technologies.
This work provides the first theoretical analysis to demonstrate that latent representations of synthetic images generated by GANs and DMs form significantly different submanifolds in feature space. This analysis motivates TriDetect, a semi-supervised detection method that enhances binary classification with architecture-aware clustering. Through discovering latent architectural patterns without explicit labels, TriDetect can learn more generalized representations that help to improve the generalization capabilities across unseen generators within the same architecture families. This work establishes a new paradigm for AI-generated image detection, shifting from pattern recognition to architectural understanding, providing a foundation for addressing future generators.

% \bigskip

\newpage
\section{Acknowledgments}
This work was funded by Taighde Éireann – Research Ireland through the Research Ireland Centre for Research Training in Machine Learning (18/CRT/6183).

\bibliography{aaai2026}

\newpage

\onecolumn
\section*{Appendix for \textit{Beyond Binary Classification: A Semi-supervised Approach to Generalized AI-generated Image Detection}}
\section{Proof of Theorem}
\subsection{Key Definitions}
\begin{definition}(GAN objective)
Let $\mathcal{D}$ and $\mathcal{G}$ be the Discriminator and Generator of GAN, respectively. The GAN objective is defined by 
\begin{equation}
    \min_{\mathcal{G}}\max_{\mathcal{D}}V(\mathcal{D},\mathcal{G}) = \mathbb{E}_{x \sim p_{\text{data}}}[\log(\mathcal{D}(x))] + \mathbb{E}_{z \sim p_{z}}[\log(1 - \mathcal{D}(\mathcal{G}(z)))],
\end{equation}
where $V(\mathcal{D},\mathcal{G})$ represents the value function, $p_z$ denotes the generator's distribution from which input vector $z$ is sampled, and $\mathcal{D}(\mathcal{G}(z))$  is the output of the Discriminator for a generated sample $\mathcal{G}(z)$.
\label{def:GAN}
\end{definition}

\begin{definition} (Diffusion Process)
    There are two processes in Diffusion Process, including the forward process and the reverse process:
    \begin{enumerate}
        \item The forward process is defined as:
        \begin{equation}
            q(x_t | x_{t-1}) = \mathcal{N}(x_t; \sqrt{1 - \beta_t}x_{t-1}, \beta_t I) \quad  \quad q(x_t | x_0) = \mathcal{N}(x_t; \sqrt{\tilde{\alpha}_t}x_0, (1 - \tilde{\alpha}_t)I),
        \end{equation}
        where $\tilde{\alpha}_t = \prod_{s=1}^{t}\alpha_s$ and $\alpha_t = 1 - \beta_t$, with $\beta \in (0,1)$ is a variance schedule. 
        \item The reverse process is defined as:
        \begin{equation}
           p_{\theta}(x_{0:T}) = p(x_T)\prod_{t=1}^{T}p_{\theta}(x_{t-1} | x_t) \quad \quad
            p_{\theta}(x_{t-1} | x_t) = \mathcal{N}(x_{t-1}; \mu_{\theta}(x_t, t), \Sigma_{\theta}(x_t, t)),
        \end{equation}
        where $\mu_{\theta}(x_t, t)$ is the mean of the reverse process distribution, which is predicted by a neural network with parameters $\theta$, $\Sigma_{\theta}(x_t, t)$ is the covariance matrix of the reverse process distribution, which is also learned, and $\theta$ represents the learnable parameters of a neural network. 
    \end{enumerate}
    \label{def:diffusion-process}
\end{definition}

\begin{definition}(Evidence Lower Bound)
    Let X, Y be random variables that have joint distribution $p_{\theta}$, and assume that $x \sim p_{\text{data}}$ then Evidence Lower Bound, denoted ELBO, defined as: 
\begin{equation}
    \text{ELBO}(x) = \mathbb{E}_{y \sim q(\cdot|x)}[\log p_{\theta}(x, y) - \log q(y|x)],
    \label{eq:ELBO}
\end{equation}
where $q$ is any distribution. 
    \label{def:ELBO}
\end{definition}

\begin{definition}(Kullback-Leibler (KL) divergence)
    The KL divergence is a measure of how different two probability distributions are. Assum that $p(x)$ is the true distribution and and $q(x)$ is the approximate distribution.  The Kullback-Leibler (KL) divergence defined by:
    \begin{equation}
        D_{KL}(P||Q) = \int p(x) \log\left(\frac{p(x)}{q(x)}\right)dx.
    \end{equation}
    \label{def:KL}
\end{definition}

\begin{definition}(Jensen-Shannon (JS) Divergence)
    Let P and Q be two probability distributions over the same probability space. The Jensen-Shannon Divergence is defined as:
    \begin{equation}
        D_{JS}(P \| Q) = \frac{1}{2}D_{KL}(P \| M) + \frac{1}{2}D_{KL}(Q \| M),
    \end{equation}
    where $M = \frac{1}{2}(P+Q)$ is the mixture distribution of $P$ and $Q$.
    \label{def:JS}
\end{definition}

\subsection{Proof of Lemmas}
\subsection{Lemma \ref{lemma:DM}}
\begin{proof}
    From Definition \ref{def:KL}, we have the KL divergence between $p_{text{data}}$ and $p_{\text{DM}}$:
    \begin{align}
        \begin{split}
        D_{KL}(p_{\text{data}}||p_{DM}) &= \int p_{\text{data}}(x_0) \log \left(\frac{p_{\text{data}}(x_0)}{p_{\text{DM}}(x_0)}\right) \,dx_0 \\
        &= \int p_{\text{data}}(x_0) \log p_{\text{data}}(x_0) dx_0 - \int p_{\text{data}}(x_0) \log p_{\text{DM}}(x_0) dx_0  \\
        &= \mathbb{E}_{x_0 \sim p_{\text{data}}}[\log p_{\text{data}}(x_0)] - \mathbb{E}_{x_0 \sim p_{\text{data}}}[\log p_{\text{DM}}(x_0)] \\
        &= -\text{H}(p_{\text{data}}) - \mathbb{E}_{\{x_0 \sim p_{\text{data}}\}}[\log p_{\text{DM}}(x_0)] \quad (\text{By Information Theory \cite{cover1999elements}}) 
        \end{split}
    \end{align}

    With $p_{\text{DM}}(x_0)$, Eq. \ref{eq:ELBO} in Definition \ref{def:ELBO}, for any distribution $q$, can be re-written as:
    \begin{equation}
    \text{ELBO}(x) = \mathbb{E}_{y \sim q(\cdot|x_0)}[\log p_{\text{DM}}(x_0, y) - \log q(y|x_0)]
    \end{equation}
    
    Let $y = x_{1:T}$ (the latent variables are the noisy versions of $x_0$), and $q(x_{1:T} | x_0)$ is the forward process from Definition \ref{def:diffusion-process}.

    By the variational inference principle:
    \begin{align}
        \begin{split}
             \log p_{\text{DM}}(x_0) &= \log \int p_{\text{DM}}(x_0, x_{1:T}) \,dx_{1:T} \\
            &= \log \int q(x_{1:T}|x_0) \cdot \left[\frac{ p_{\text{DM}}(x_0, x_{1:T})}{q(x_{1:T}|x_0)}\right] \,dx_{1:T} \\
            &\ge \int q(x_{1:T}|x_0) \log \left[\frac{p_{\text{DM}}(x_0, x_{1:T})}{q(x_{1:T}|x_0)}\right] \,dx_{1:T} \quad \text{(by Jensen's inequality)} \\
            &= \text{E}_{x_{1:T} \sim q(\cdot|x_0)}[\log p_{\text{DM}}(x_0, x_{1:T}) - \log q(x_{1:T}|x_0)] \\
            &= \text{ELBO}(x_0)
        \end{split}
    \end{align}

    Therefore: 
    \begin{equation}
        \log p_{\text{DM}}(x_0) \ge \text{ELBO}(x_0)
    \end{equation}
    which implies:
    \begin{equation}
        -\log p_{\text{DM}}(x_0) \le -\text{ELBO}(x_0)
    \end{equation}
    Substituting back into our expression for $D_{KL}$:
    \begin{align}
    \begin{split}
        D_{KL}(p_{\text{data}}||p_{\text{DM}}) &= -\text{H}(p_{\text{data}}) - \mathbb{E}_{\{x_0 \sim p_{\text{data}}\}}[\log p_{\text{DM}}(x_0)] \\
        &\le -\text{H}(p_{\text{data}}) + \mathbb{E}_{\{x_0 \sim p_{\text{data}}\}}[-\text{ELBO}(x_0)]
    \end{split}
    \end{align}

    Therefore, minimizing $\mathbb{E}_{\{x_0 \sim p_{\text{data}}\}}[-\text{ELBO}(x_0)]$ provides an upper bound minimization for $D_{\text{KL}}(p_{\text{data}}||p_{\text{DM}})$.
\end{proof}

\subsubsection{Lemma \ref{lemma:GAN}}
\begin{proof}
    For a fixed generator $\mathcal{G}$ and any discriminator $\mathcal{D}$, we have (from Definition \ref{def:GAN}):
    \begin{equation}
        V(\mathcal{D}, \mathcal{G}) = \mathbb{E}_{x \sim p_{\text{data}}}[\log(\mathcal{D}(x))] + \mathbb{E}_{z \sim p_{z}}[\log(1 - \mathcal{D}(\mathcal{G}(z)))]
    \end{equation}

    Since $\mathcal{G}(z)$ with $z \sim p_z$ generates samples according to the distribution $p_{\text{GAN}}$, we can rewrite:
    \begin{equation}
        V(\mathcal{D}, \mathcal{G}) = \mathbb{E}_{x \sim p_{\text{data}}}[\log(\mathcal{D}(x))] + \mathbb{E}_{x \sim p_{\text{GAN}}}[\log(1 - \mathcal{D}(x))]
    \end{equation}

    This can be expressed as an integral:
    \begin{align}
        \begin{split}
             V(\mathcal{D}, \mathcal{G}) &= \int p_{\text{data}}(x) \log(\mathcal{D}(x))dx + \int p_{\text{GAN}}(x) \log(1 - \mathcal{D}(x))dx \\
        &= \int [p_{\text{data}}(x) \log(\mathcal{D}(x)) + p_{\text{GAN}}(x) \log(1 - \mathcal{D}(x))] \,dx
        \end{split}
    \end{align}

    For any fixed $x$, to maximize the integrated $f(D(x)) = p_{data}(x) \log D(x) + p_{GAN}(x) \log(1 - D(x))$ with respect to $\mathcal{D}(x) \in (0,1)$, we take the derivative:
    \begin{equation}
       f'(\mathcal{D}) = \frac{p_{\text{data}}(x)}{\mathcal{D}} - \frac{p_{\text{GAN}}(x)}{1-\mathcal{D}}
    \end{equation}
    Setting $f'(\mathcal{D}) = 0$ and solving:
    \begin{align}
        \begin{split}
            \frac{p_{\text{data}}(x)}{\mathcal{D}(x)} &= \frac{p_{\text{GAN}}(x)}{1-\mathcal{D}(x)} \\
        p_{\text{data}}(x)(1 - \mathcal{D}(x)) &= p_{\text{GAN}}(x)\mathcal{D}(x) \\
        p_{\text{data}}(x) &= \mathcal{D}(x)(p_{\text{data}}(x) + p_{\text{GAN}}(x))
        \end{split}
    \end{align}

    Therefore, the optimal discriminator:
    \begin{equation}
        \mathcal{D}^*(x) = \frac{p_{\text{data}}(x)}{p_{\text{data}}(x) + p_{\text{GAN}}(x)}
    \end{equation}

    To verify this is a maximum, check the second derivative:
    \begin{equation}
        f''(\mathcal{D}) = -\frac{p_{\text{data}}(x)}{\mathcal{D}^2} - \frac{p_{\text{GAN}}(x)}{(1-\mathcal{D})^2} < 0
    \end{equation}
    Since $f''(\mathcal{D}) < 0$, $\mathcal{D}$ gives the maximum value. As we know $\mathcal{D}^*(x)$ maximizes the the integrand for each $x$, we have for any discriminator $\mathcal{D}$:
    \begin{equation}
        p_{\text{data}}(x) \log D(x) + p_{\text{GAN}}(x) \log(1 - D(x)) \le p_{\text{data}}(x) \log D^{*}(x) + p_{\text{GAN}}(x) \log(1 - D^{*}(x))
    \end{equation}
    Integrating over all $x$:
    \begin{equation}
        V(\mathcal{D}, \mathcal{G}) \leq V(\mathcal{D}^*, \mathcal{G}),
        \label{eq:inequality-GAN}
    \end{equation}
    with equality if and only if $\mathcal{D}(x) = \mathcal{D}^*(x)$ for almost all $x$.
    
    Next, we substitute $D^*(x)$ back into $V(\mathcal{D}^*, \mathcal{G})$:
    \begin{equation}
        V(\mathcal{D}^*, \mathcal{G}) = \mathbb{E}_{x \sim p_{\text{data}}} \left[ \log \frac{p_{\text{data}}(x)}{p_{\text{data}}(x) + p_{\text{GAN}}(x)} \right] + \mathbb{E}_{x \sim p_{\text{GAN}}} \left[ \log \frac{p_{\text{GAN}}(x)}{p_{\text{data}}(x) + p_{\text{GAN}}(x)} \right]
    \end{equation}

    Recall from Definition \ref{def:JS}, for our case, let $M = \frac{1}{2}(p_{data} + p_{\text{GAN}})$. Then:
    \begin{align}
        \begin{split}
             D_{JS}(p_{\text{data}}\|p_{\text{GAN}}) &= \frac{1}{2}D_{KL}(p_{\text{data}} \| M) + \frac{1}{2}D_{KL}(p_{\text{GAN}} \| M) \\
        &=  \frac{1}{2}\mathbb{E}_{x \sim p_{\text{data}}} \left[ \log \frac{p_{\text{data}}(x)}{M(x)} \right] + \frac{1}{2}\mathbb{E}_{x \sim p_{\text{GAN}}} \left[ \log \frac{p_{\text{GAN}}(x)}{M(x)} \right]\\
        &= \frac{1}{2}\mathbb{E}_{x \sim p_{\text{data}}} \left[ \log \frac{2p_{\text{data}}(x)}{p_{\text{data}}(x)+p_{\text{GAN}}(x)} \right] + \frac{1}{2}\mathbb{E}_{x \sim p_{\text{GAN}}} \left[ \log \frac{2p_{\text{GAN}}(x)}{p_{\text{data}}(x)+p_{\text{GAN}}(x)} \right]
        \end{split} 
    \end{align}
    
    Using the property $\log(2a) = \log 2 + \log a$:
    \begin{align}
        \begin{split}
            D_{JS}(p_{\text{data}} \| p_{\text{GAN}}) &= \frac{1}{2} \log 2 + \frac{1}{2}\mathbb{E}_{x \sim p_{\text{data}}} \left[ \log \frac{p_{\text{data}}(x)}{p_{\text{data}}(x)+p_{\text{GAN}}(x)} \right] \\
        &+ \frac{1}{2} \log 2 + \frac{1}{2}\mathbb{E}_{x \sim p_{\text{GAN}}} \left[ \log \frac{p_{\text{GAN}}(x)}{p_{\text{data}}(x)+p_{\text{GAN}}(x)} \right] \\
        &= \log 2 + \frac{1}{2}V(\mathcal{D}^*, \mathcal{G})
        \end{split}
    \end{align}

    Rearranging, we have:
    \begin{equation}
        V(\mathcal{D}^*, \mathcal{G}) = 2D_{JS}(p_{\text{data}} \| p_{\text{GAN}}) - 2\log2
        \label{eq:GAN-optimal}
    \end{equation}

    Combining Eq. \ref{eq:inequality-GAN} and \ref{eq:GAN-optimal}, we have:
    \begin{equation}
        V(\mathcal{D}, \mathcal{G}) \le V(\mathcal{D}^*, \mathcal{G} = 2D_{JS}(p_{\text{data}} || p_{\text{GAN}}) - 2 \log 2
    \end{equation}
\end{proof}

\subsection{Complete Proof of Theorem \ref{theorem:difference-optimization}}
\begin{proof}
\textbf{Part 1: GAN Optimization:} \\

    From Lemma \ref{lemma:GAN}, we established that for a fixed generator $\mathcal{G}$, the value function satisfies:
    \begin{equation}
         V(\mathcal{D}, \mathcal{G}) \leq 2D_{\text{JS}}(p_{\text{data}} \| p_{\text{GAN}}) - 2\log2
    \end{equation}

    When the discriminator is optimal, we have:
    \begin{equation}
        V(\mathcal{D}^*, \mathcal{G}) = 2D_{\text{JS}}(p_{\text{data}} \| p_{\text{GAN}}) - 2\log2
    \end{equation}

    The GAN training objective for the generator is is to minimize the value function with respect to $\mathcal{G}$:
    \begin{equation*}
        \min_{\mathcal{G}} V(\mathcal{D}^*, \mathcal{G}) = \min_{\mathcal{G}}[2D_{JS}(p_{\text{data}}\|p_{\text{GAN}}) - 2\log2]
    \end{equation*}

    Since $-2\log 2$ is a constant independent of $\mathcal{G}$, minimizing $V(\mathcal{D}^*, \mathcal{G})$ is equivalent to minimizing $D_{\text{JS}}(p_{\text{data}}|p_{\text{GAN}})$:
    \begin{equation}
         \min_{\mathcal{G}} V(\mathcal{D}^*, \mathcal{G}) \equiv \min D_{\text{JS}}(p_{\text{data}}\|p_{\text{GAN}})
    \end{equation}
    Therefore, GANs minimize the JS divergence between the data distribution and the generated distribution.

\textbf{Part 2: DM Optimization:}

    From Lemma \ref{lemma:DM}, we established that DMs minimize an upper bound on the KL divergence:
    \begin{equation}
        D_{KL}(p_{\text{data}}||p_{\text{DM}}) \leq \mathbb{E}_{\{x_0 \sim p_{\text{data}}\}}[-\text{ELBO}(x_0)] + \text{H}(p_{\text{data}})
    \end{equation}
    
    The DM training objective is to maximize the ELBO, which is equivalent to minimizing $-ELBO$:
    \begin{equation}
        \min_{\theta} \mathbb{E}_{x_0 \sim p_{\text{data}}}[-\text{ELBO}(x_0)],
    \end{equation}
    where $\theta$ represents the parameters of the reverse process.

    From Remark \ref{remark:DM}, the inequality becomes an equality when the variational posterior   $q(x_{1:T}|x_0)$ (the forward process) matches the true posterior $p_{\theta}(x_{1:T}|x_0)$ under the model. At the global optimum with sufficient model capacity:

    $$D_{KL}(p_{\text{data}} \| p_{\text{DM}}) = \mathbb{E}_{x_0 \sim p_{\text{data}}}[-\text{ELBO}(x_0)] + H(p_{\text{data}})$$

    Since $H(p_{\text{data}})$ is a constant (independent of model parameters), minimizing the right-hand side is equivalent to minimizing:
    \begin{equation}
        \min_{\theta} \mathbb{E}_{x \sim p_{\text{data}}}[-\text{ELBO}(x)] \equiv \min_{\theta} D_{KL}(p_{\text{data}} \| p_{\text{DM}})
    \end{equation}
\end{proof}

\subsection{Complete Proof of Theorem \ref{theorem:distribution-support}}
We assume all distributions admit densities with respect to a common base measure $\mu$ on $\mathcal{X}$. For notational convenience, we use the same symbols for distributions and their densities.

\begin{proof}
    \textbf{Part 1: GANs Support Analysis.} From Definition \ref{def:JS}, the JS divergence between $p_{\text{data}}$ and $p_{GAN}$ is defined as:
    \begin{align}
        \begin{split}
            D_{JS}(p_{\text{data}} \|p_{\text{GAN}}) &= \frac{1}{2}D_{KL}(p_{\text{data}} \| M)   + \frac{1}{2}D_{KL}(p_{\text{GAN}} \| M) \\
        &= \frac{1}{2}\int \underbrace{p_{\text{data}}(x)\log \left(\frac{p_{\text{data}}(x)}{M(x)}\right) dx}_{(*)} + \frac{1}{2} \int \underbrace{p_{\text{GAN}}(x) \log \left(\frac{p_{\text{GAN}}(x)}{M(x)}\right) dx}_{(**)}
        \end{split}
        \label{eq:JS}
    \end{align}
    where $M = \frac{1}{2}(p_{\text{data}} + p_{\text{GAN}})$  is the mixture distribution.

    Let us analyze the first integral for any $x \in \mathcal{S}_{\text{data}}$ (where $p_{\text{data}}(x) > 0$):
    \begin{itemize}
        \item \textit{Case 1:} If $p_{\text{GAN}}(x) = 0$, then $M(x) = \frac{1}{2}p_{\text{data}}(x) > 0$. The contribution to the first term (*) at point $x$ is:
        \begin{equation}
            \lim_{p_{\text{GAN}}(x) \to 0} p_{\text{data}}(x)\log \left(\frac{p_{\text{data}}(x)}{M(x)}\right) = p_{\text{data}}(x)\log \left(\frac{p_{\text{data}}(x)}{\frac{1}{2}p_{\text{data}}(x)}\right) = p_{\text{data}}(x)\log 2
        \end{equation}
        Since $\log 2$ is finite, the integral of the first term (*) $\int p_{\text{data}}(x)\log \left(\frac{p_{\text{data}}(x)}{M(x)}\right) < 0$ remains finite even when $S_{\text{GAN}} \subset S_{\text{data}}$.
        
        \item \textit{Case 2:} If $p_{\text{GAN}}(x) > 0$, then $M(x) = \frac{1}{2}(p_{\text{data}} + p_{\text{GAN}}) > 0$ and the integral of the first term (*) is finite. 

        For the second KL term (**), we only integrate over ${x : p_{GAN}(x) > 0}$. Since $p_{GAN}(x) > 0 \implies M(x) \geq \frac{1}{2}p_{GAN}(x) > 0$, this integral is well-defined and finite.
    
        Therefore:
        \begin{equation}
            D_{JS}(p_{data}||p_{GAN}) = \frac{1}{2}D_{KL}(p_{data}||M) + \frac{1}{2}D_{KL}(p_{GAN}||M) < \infty        
        \end{equation}
    
        The JS divergence remains finite even when $S_{GAN} \subset S_{data}$, allowing GANs to achieve optimal solutions without full support coverage.
    \end{itemize}
    
    \textbf{Part 2: DMs Support Analysis}

    From Lemma \ref{lemma:DM}, DMs minimize:
    \begin{equation}
        D_{KL}(p_{\text{data}} \| p_{\text{DM}}) = \int p_{\text{data}}(x) \log\left(\frac{p_{\text{data}}(x)}{p_{\text{DM}}(x)}\right) dx
    \end{equation}

    Suppose there exists a set $\mathcal{A} \subseteq S_{\text{data}}$ with positive measure under $p_{\text{data}}$ such that $p_{\text{DM}}(x) = 0$ for all $x \in \mathcal{A}$. We can decompose the integral: 
    \begin{equation}
        D_{KL}(p_{\text{data}}|p_{\text{DM}}) = \underbrace{\int_\mathcal{A} p_{\text{data}}(x) \log\left(\frac{p_{\text{data}}(x)}{p_{\text{DM}}(x)}\right)dx}_{\text{Problematic part: } p_{\text{DM}}(x)=0} + 
        \underbrace{\int_{\mathcal{S}_{\text{data}} \setminus \mathcal{A}} p_{\text{data}}(x) \log\left(\frac{p_{\text{data}}(x)}{p_{\text{DM}}(x)}\right)dx}_{\text{Well-behaved part: } p_{\text{DM}}(x)>0}
    \end{equation}

    For any $x \in \mathcal{A}$, we have $p_{\text{data}}(x) > 0$ (since $x \in \mathcal{S}{\text{data}}$) and $p{DM}(x) = 0$. To evaluate the integral over $\mathcal{A}$, we consider the limit:
    \begin{equation}
        \lim_{p_{\text{DM}}(x) \to 0^+} p_{\text{data}}(x) \log \left( \frac{p_{\text{data}}(x)}{p_{\text{DM}}(x)} \right) = p_{\text{data}}(x) \cdot \lim_{p_{\text{DM}}(x) \to 0^+} \log \left( \frac{p_{\text{data}}(x)}{p_{\text{DM}}(x)} \right) = +\infty
    \end{equation}

    Since $p_{\text{data}}(x) > 0$ is fixed and $\lim_{p_{\text{DM}}(x) \to 0^+} \log\left(\frac{p_{\text{data}}(x)}{p_{\text{DM}}(x)}\right) = +\infty$.

    Therefore:
    \begin{equation}
        \int_A p_{\text{data}}(x) \log\left(\frac{p_{\text{data}}(x)}{p_{\text{DM}}(x)}\right) dx = +\infty
    \end{equation}

    This implies:
    \begin{equation}
        D_{\text{KL}}(p_{\text{data}}||p_{\text{DM}}) = +\infty
    \end{equation}

    Since the optimization objective is to minimize $D_{\text{KL}}(p_{\text{data}}\|p_{\text{DM}})$, any distribution $p_{\text{DM}}$ with $S_{\text{DM}} \not\supseteq \mathcal{S}_{\text{data}}$ yields an infinite objective value and cannot be optimal. For DMs to achieve a finite (and thus optimizable) KL divergence, we must have $\mathcal{S}_{\text{data}} \subseteq \mathcal{S}_{\text{DM}}$. 
\end{proof}

\section{Sinkhorn-Knopp Algorithm}
\begin{algorithm}
\caption{Sinkhorn-Knopp Algorithm}
\label{alg:sinkhorn}
\begin{algorithmic}[1]
\Require Logits $s^{\text{fake}} \in \mathbb{R}^{B \times K}$, temperature $\epsilon$, iterations $T$
\Ensure Balanced assignment matrix $Q \in \mathbb{R}^{B \times K}$
\State $Q^{(0)} \gets \exp(s^{\text{fake}} / \epsilon)$ \Comment{Initialize with softmax}
\For{$t = 1$ to $T$}
    \State $Q^{(t)} \gets \mathcal{R}(Q^{(t-1)})$ \Comment{Row normalization}
    \State $Q^{(t)} \gets \mathcal{C}(Q^{(t)})$ \Comment{Column normalization}
\EndFor
\State $Q \gets \mathcal{R}(Q^{(T)})$ \Comment{Final row normalization}
\State \Return $Q$
\end{algorithmic}
\end{algorithm}

\section{Experimental Details}
\subsection{Datasets}
We evaluate TriDetect on 2 standard benchmarks (GenImage \cite{zhu2023genimage}, AIGCDetectBenchmark \cite{zhong2023patchcraft}) and 3 in-the-wild datasets (WildFake \cite{hong2025wildfake}, Chameleon \cite{yan2025a}, DF40 \cite{yan2024df40}):

\begin{itemize}
    \item GenImage \cite{zhu2023genimage}: GenImage includes 331,167 real images and 1,350,000 AI-generated images. Synthetic images generated by generators: Midjourney \cite{midjourneyMidjourney}, SD (V1.4 and V1.5) \cite{rombach2022high}, ADM \cite{dhariwal2021diffusion}, GLIDE \cite{nichol2021glide}, Wukong \cite{mindsporex6607x601Dx5927x6A21x578Bx5E73x53F0}, VQDM \cite{gu2022vector}, and BigGAN \cite{brock2018large}.
    \item AIGCDetectBenchmark \cite{zhong2023patchcraft}: This dataset includes synthetic images created by 16 generators: ProGAN \cite{karras2017progressive}, StyleGAN \cite{karras2019style}, BigGAN \cite{brock2018large}, CycleGAN \cite{zhu2017unpaired}, StarGAN \cite{choi2018stargan}, GauGAN \cite{park2019semantic}, Stylegan2 \cite{karras2020analyzing}, ADM \cite{dhariwal2021diffusion}, Glide GLIDE \cite{nichol2021glide}, VQDM \cite{gu2022vector}, Wukong \cite{mindsporex6607x601Dx5927x6A21x578Bx5E73x53F0}, Midjourney \cite{midjourneyMidjourney}, Stable Diffusion (SDv1.4, SDv1.5) \cite{rombach2022high}, DALL-E 2 \cite{ramesh2022hierarchical}, and SD\_XL \cite{rombach2022high}.
    \item WildFake \cite{hong2025wildfake}: Testing sets in this dataset applied by a series of degradation techniques: downsampling, JPEG compression, geometric transformations (flipping, cropping), adding watermarks (textual or visual), and color transformations. We evaluate TriDetect on these generators: DALL-E \cite{ramesh2022hierarchical}, DDPM \cite{nichol2021improved}, DDIM \cite{song2020denoising}, VQDM \cite{gu2022vector}, BigGAN \cite{brock2018large}, StarGAN \cite{choi2018stargan}, StyleGAN \cite{karras2019style}, DF-GAN \cite{tao2022df}, GALIP \cite{tao2023galip}, GigaGAN \cite{kang2023scaling}. 
    \item DF40 \cite{yan2024df40}: DF40 contains more than 40 GAN-based and DM-based generators. In our work, we select 7 generators for evaluation: CollabDiff \cite{huang2023collaborative}, MidJourney \cite{midjourneyMidjourney}, DeepFaceLab \cite{githubGitHubIperovDeepFaceLab}, StarGAN \cite{choi2018stargan}, StarGAN2 \cite{choi2020stargan}, StyleCLIP \cite{patashnik2021styleclip}, and WhichisReal \cite{whichfaceisrealWhichFace}. 
    \item Chameleon \cite{yan2025a}: Synthetic images in this dataset are designed to be challenging for human perception. The dataset contains approximately 26,000 test images in total, without separated into different generator subsets. 
\end{itemize}

\subsection{Implementation}

Our implementation employs a pre-trained CLIP ViT-L/14 vision encoder ($f$) with 1024-dimensional output features, then fine-tuned using Low-Rank Adaptation (LoRA) with rank $r=16$ and scaling factor $\alpha=32$, applied exclusively to the query and key projection matrices. The architecture-aware classifier consists of a three-layer MLP classifier $g_\theta: \mathbb{R}^{1024} \rightarrow \mathbb{R}^3$ with hidden dimensions [256, 128] and ReLU activations, producing logits for one real class and two fake clusters. The Sinkhorn-Knopp algorithm operates with temperature $\epsilon=0.05$ and $T=3$ iterations, ensuring balanced cluster assignments where each cluster receives approximately $B/K$ samples. 

For training stability, we compute binary logits using the numerically stable log-sum-exp formulation: $Z_{\text{real}} = z_0$ and $Z_{\text{fake}} = \log(\exp(z_1) + \exp(z_2))$. 
The cross-prediction loss is implemented by computing balanced assignments $Q = \text{Sinkhorn}(Z_{\text{fake}})$ and $Q' = \text{Sinkhorn}(Z'_{\text{fake}})$ for original and augmented views respectively, then optimizing  $\mathcal{L}{\text{cross}} = -\frac{1}{2}[\sum_i Q'i \log p_i + \sum_i Q_i \log p'i]$ with detached assignments to prevent trivial solutions. 

We use Adam optimizer with learning rate $\eta=2 \times 10^{-4}$, $(\beta_1, \beta_2) = (0.9, 0.95)$, and weight decay $\lambda=10^{-4}$. The total loss combines binary classification and clustering objectives with weights $\mathcal{L}_{\text{total}} = 0.7\mathcal{L}_{\text{binary}} + 0.3\mathcal{L}_{\text{cluster}}$, where the cluster loss includes cross-prediction ($\omega_1=1.0$) and cross-view consistency ($\omega_2=0.1$) terms. 
Training is conducted with batch size $B=128$ for $5$ epochs on randomly shuffled GAN and diffusion-generated images. CLIP-specific normalization is applied to $224 \times 224$ input images, where each RGB channel is normalized as $x' = \frac{x - \mu}{\sigma}$ with per-channel means $\boldsymbol{\mu} = [0.481,\, 0.458,\, 0.408]$ and standard deviations $\boldsymbol{\sigma} = [0.269,\, 0.261,\, 0.276]$.

For CNNSpot \cite{wang2020cnn}, LNP \cite{liu2022detecting}, FreDect \cite{frank2020leveraging}, Fusing \cite{ju2022fusing}, LGrad \cite{tan2023learning}, DIRE \cite{wang2023dire}, UnivFD \cite{ojha2023towards}, AIDE \cite{yan2025a} baselines, we utilize the implementation provided by AIGCDetectBenchmark \cite{zhong2023patchcraft} for implementation. Regarding CORE \cite{ni2022core}, SPSL \cite{liu2021spatial}, UIA-ViT \cite{zhuang2022uia} and Effort \cite{yan2024orthogonal}, we use the implementation provided by DeepfakeBench \cite{yan2023deepfakebench}. Excluding from NPR \cite{tan2024rethinking}, we use the official source code provided by the authors. 

All experiments are conducted using a single run with a fixed random seed of 1024. To ensure robust evaluation despite using a single run, we compensate by evaluating each method across multiple diverse benchmarks. Our computational environment consists of an NVIDIA H100 NVL GPU with 94GB of VRAM, 48 CPU cores on a Linux operating system. The implementation is based on PyTorch 2.3.1 with Python 3.9.

\begin{table}[ht]
\centering
\fontsize{8pt}{8pt}\selectfont
\caption{Performance evaluation on Chameleon dataset with existing attribution baselines.}
\label{table:compare-attribution}
\begin{tblr}{
  width = \linewidth,
  colspec = {Q[508]Q[106]Q[106]Q[106]Q[106]},
  cells = {c},
  vline{2-5} = {-}{},
  hline{1,5} = {-}{0.08em},
  hline{2,4} = {-}{},
  hline{2} = {2}{-}{},
}
                                    & \textbf{AUC}    & \textbf{ACC}    & \textbf{EER}    & \textbf{AP}     \\
Cross-entropy loss                  & 0.7265          & 0.5963          & 0.3411          & 0.6493          \\
ArcFace loss \cite{deng2019arcface} & 0.7616          & 0.6104          & 0.3007          & 0.7017          \\
Ours                                & \textbf{0.8935} & \textbf{0.6788} & \textbf{0.1843} & \textbf{0.8742} 
\end{tblr}
\end{table}

\begin{table}[ht]
\centering
\fontsize{8pt}{8pt}\selectfont
\caption{Ablation studies on different numbers of fake clusters. Results are evaluated on Chameleon dataset.}
\label{table:clusters}
\begin{tblr}{
  width = \linewidth,
  colspec = {Q[423]Q[125]Q[125]Q[125]Q[125]},
  cells = {c},
  vline{2} = {-}{},
  hline{1,6} = {-}{0.08em},
  hline{2} = {-}{},
  hline{2} = {2}{-}{},
}
\textbf{Number of fake clusters} & \textbf{AUC}    & \textbf{ACC}    & \textbf{EER}    & \textbf{AP}     \\
2                                & \textbf{0.8935} & \textbf{0.6788} & \textbf{0.1843} & \textbf{0.8742} \\
3                                & 0.8191          & 0.5978          & 0.2538          & 0.7847          \\
4                                & 0.8211          & 0.5867          & 0.2518          & 0.7756          \\
5                                & 0.7396          & 0.5911          & 0.3246          & 0.6979          
\end{tblr}
\end{table}

\section{Additional Ablations and Results}
In this section, we provide additional experimental results and ablation studies of our proposed method. 

\subsection{Comparison with Attribution Baselines}
Table \ref{table:compare-attribution} presents a comparison between our semi-supervised detection approach and existing attribution baselines. It is important to emphasize that our method is not designed for the attribution task, which identifies which specific generator produced a fake image. Instead, our approach focuses on discovering latent architectural patterns to enhance binary detection performance. Despite this fundamental difference in objectives, we include this comparison to demonstrate the effectiveness of our approach. The results reveal that our method substantially outperforms both attribution baselines, showing that discovering and leveraging architectural patterns through semi-supervised learning is effective for generalized detection.

\subsection{Ablation Study on Different Number of Fake Clusters}
Table \ref{table:clusters} demonstrates that the optimal number of fake clusters is $K=2$, which achieves the best performance across all metrics with an AUC of 0.8935 and the lowest EER of 0.1843. This result strongly aligns with our objective of having the model discover distinct architectural patterns that separate GAN-generated images from those created by diffusion models in the feature space.

\subsection{Experimental Results on Standard Benchmarks and In-the-wild Datasets}
In our manuscript, we mainly present the AUC results on GenImage \cite{zhu2023genimage} and AIGCDetectBenchmark \cite{zhong2023patchcraft} datasets. This section reports the evaluation results on these two datasets across 3 metrics: ACC, EER, and AP. On GenImage (Tables \ref{table:GenImage-ACC}-\ref{table:GenImage-AP}), TriDetect achieves an average ACC of 0.9111, the lowest EER of 0.0343, and the average precision of 0.9894. On AIGCBenchmark (Tables \ref{table:AIGC-ACC}-\ref{table:AIGC-AP}), TriDetect also outperforms SoTA methods. 

Regarding challenging real-world datasets: WildFake \cite{hong2025wildfake} and DF40 \cite{yan2024df40}, only ACC results are reported in the main manuscript. In this section, tables \ref{table:wildfake-ACC}-\ref{table:DF40-AP} demonstrate TriDetect's generalization capability across other metrics: AUC, EER, and AP. Although degradation techniques are applied to testing sets of WildFake, our method shows higher results than other methods across all metrics. This demonstrates the better robustness of TriDetect than other methods.

\begin{table*}[ht]
\centering
\fontsize{7pt}{7pt}\selectfont
\caption{Comparison on the GenImage \cite{zhu2023genimage} Dataset in terms of ACC Performance. The best result and the second-best result are marked in \textbf{bold} and \ul{underline}, respectively.}
\label{table:GenImage-ACC}
\begin{tblr}{
  width = \linewidth,
  colspec = {Q[130]Q[104]Q[100]Q[87]Q[87]Q[100]Q[100]Q[100]Q[100]},
  cells = {c},
  row{15} = {FrenchPass},
  vline{2-9} = {-}{},
  hline{1,16} = {-}{0.08em},
  hline{2,15} = {-}{},
  hline{2} = {2}{-}{},
}
\textbf{Method} & \textbf{BigGAN} & \textbf{SD v1.4} & \textbf{ADM}    & \textbf{GLIDE}  & \textbf{MidJourney} & \textbf{VQDM}   & \textbf{Wukong} & \textbf{Avg}    \\
CNNSpot         & 0.9988          & 0.9910           & 0.5170          & 0.5882          & 0.5344              & 0.5106          & 0.9658          & 0.7294          \\
FreDect         & 0.9853          & 0.9720           & 0.5368          & 0.5533          & 0.5473              & 0.6113          & 0.9443          & 0.7358          \\
Fusing          & 0.9988          & 0.9948           & 0.5081          & 0.5664          & 0.5098              & 0.5317          & 0.9718          & 0.7259          \\
LGrad           & 0.6037          & 0.5308           & 0.5181          & 0.6103          & 0.5301              & 0.5181          & 0.5588          & 0.5528          \\
LNP             & 0.9138          & 0.9178           & 0.5056          & 0.6089          & 0.5362              & 0.5398          & 0.8878          & 0.7014          \\
CORE            & 0.9986          & 0.9933           & 0.5683          & 0.9498          & 0.5299              & 0.5580          & 0.9643          & 0.7946          \\
SPSL            & 0.9940          & 0.9313           & 0.5008          & 0.5358          & 0.5292              & 0.5192          & 0.8925          & 0.7004          \\
UIA-ViT         & \textbf{0.9965} & 0.9860           & 0.5038          & 0.7339          & 0.5768              & 0.5452          & 0.9322          & 0.7535          \\
DIRE            & 0.9923          & \ul{0.9943}   & 0.5988          & 0.8976          & 0.6320              & 0.6056          & 0.9713          & 0.8131          \\
UnivFD          & 0.9833          & 0.9653           & 0.7212          & \ul{0.9482}  & \ul{0.7583}      & 0.9143          & 0.8569          & \ul{0.8696}  \\
AIDE            & 0.9813          & 0.9508           & 0.5637          & 0.7567          & 0.7174              & 0.6528          & 0.8972          & 0.7885          \\
NPR             & 0.9918          & 0.9792           & \textbf{0.7745} & 0.7751          & \textbf{0.7748}     & 0.8078          & 0.9177          & 0.8601          \\
Effort          & \ul{0.9991}  & \textbf{0.9993}  & 0.5872          & 0.7942          & 0.7411              & \ul{0.9194}  & \ul{0.9878}  & 0.8612          \\
TriDetect             & \ul{0.9991}  & \textbf{0.9993}  & \ul{0.7482}  & \textbf{0.9488} & 0.7180              & \textbf{0.9679} & \textbf{0.9963} & \textbf{0.9111} 
\end{tblr}
\end{table*}

\begin{table}[ht]
\centering
\fontsize{7pt}{7pt}\selectfont
\caption{Comparison on the GenImage \cite{zhu2023genimage} Dataset in terms of EER Performance. The best result and the second-best result are marked in \textbf{bold} and \ul{underline}, respectively.}
\label{table:GenImage-EER}
\begin{tblr}{
  width = \linewidth,
  colspec = {Q[130]Q[104]Q[100]Q[87]Q[87]Q[100]Q[100]Q[100]Q[100]},
  cells = {c},
  row{15} = {FrenchPass},
  vline{2-9} = {-}{},
  hline{1,16} = {-}{0.08em},
  hline{2,15} = {-}{},
  hline{2} = {2}{-}{},
}
\textbf{Method} & \textbf{BigGAN} & \textbf{SD v1.4} & \textbf{ADM}    & \textbf{GLIDE}  & \textbf{MidJourney} & \textbf{VQDM}   & \textbf{Wukong} & \textbf{Avg}    \\
CNNSpot         & 0.0007          & 0.0052           & 0.2593          & 0.1050          & 0.2815              & 0.2747          & 0.0153          & 0.1345          \\
FreDect         & 0.0008          & 0.0275           & 0.3145          & 0.3017          & 0.3275              & 0.2697          & 0.0525          & 0.1849          \\
Fusing          & 0.0007          & 0.0045           & 0.3152          & 0.1183          & 0.3637              & 0.2898          & 0.0128          & 0.1579          \\
LGrad           & 0.4022          & 0.4673           & 0.4823          & 0.3818          & 0.4613              & 0.4827          & 0.4337          & 0.4445          \\
LNP             & 0.0703          & 0.0693           & 0.4257          & 0.2360          & 0.3890              & 0.3683          & 0.0923          & 0.2359          \\
CORE            & \ul{0.0005}  & \ul{0.0050}   & 0.2568          & 0.0273          & 0.2628              & 0.2570          & 0.0202          & 0.1185          \\
SPSL            & 0.0055          & 0.0267           & 0.4323          & 0.2482          & 0.3167              & 0.2935          & 0.0430          & 0.1951          \\
UIA-ViT         & 0.0030          & 0.0117           & 0.3313          & 0.1227          & 0.2813              & 0.2822          & 0.0348          & 0.1524          \\
DIRE            & 0.0073          & 0.0053           & 0.2518          & 0.0632          & 0.2383              & 0.2542          & 0.0253          & 0.1208          \\
UnivFD          & 0.0162          & 0.0317           & \ul{0.1527}  & 0.0433          & 0.1528              & 0.0593          & 0.0922          & 0.0783          \\
AIDE            & 0.0050          & 0.0493           & 0.2968          & 0.1463          & 0.2127              & 0.2067          & 0.0882          & 0.1436          \\
NPR             & 0.0052          & 0.0183           & 0.1462          & \ul{0.0260}  & 0.1302              & 0.1105          & 0.0610          & 0.0710          \\
Effort          & \textbf{0.0002} & \textbf{0.0007}  & 0.1763          & 0.0520          & \textbf{0.0768}     & \ul{0.0253}  & \ul{0.0067}  & \ul{0.0483}  \\
TriDetect             & \ul{0.0005}  & \textbf{0.0007}  & \textbf{0.1032} & \textbf{0.0087} & \ul{0.1133}      & \textbf{0.0110} & \textbf{0.0030} & \textbf{0.0343} 
\end{tblr}
\end{table}

\begin{table}[ht]
\centering
\fontsize{7pt}{7pt}\selectfont
\caption{Comparison on the GenImage \cite{zhu2023genimage} Dataset in terms of AP Performance. The best result and the second-best result are marked in \textbf{bold} and \ul{underline}, respectively.}
\label{table:GenImage-AP}
\begin{tblr}{
  width = \linewidth,
  colspec = {Q[130]Q[104]Q[100]Q[87]Q[87]Q[100]Q[100]Q[100]Q[100]},
  cells = {c},
  row{15} = {FrenchPass},
  vline{2-9} = {-}{},
  hline{1,16} = {-}{0.08em},
  hline{2,15} = {-}{},
  hline{2} = {2}{-}{},
}
\textbf{Method} & \textbf{BigGAN} & \textbf{SD v1.4} & \textbf{ADM}    & \textbf{GLIDE}  & \textbf{MidJourney} & \textbf{VQDM}   & \textbf{Wukong} & \textbf{Avg}    \\
CNNSpot         & \textbf{1.0000} & \ul{0.9999}   & 0.8075          & 0.9518          & 0.8025              & 0.7847          & 0.9989          & 0.9064          \\
FreDect         & \textbf{1.0000} & 0.9971           & 0.7140          & 0.7391          & 0.7193              & 0.8049          & 0.9902          & 0.8521          \\
Fusing          & \ul{0.9999}  & 0.9999           & 0.7287          & 0.9333          & 0.6845              & 0.7767          & 0.9987          & 0.8745          \\
LGrad           & 0.6380          & 0.5928           & 0.5157          & 0.6205          & 0.5417              & 0.5136          & 0.6337          & 0.5794          \\
LNP             & 0.9801          & 0.9818           & 0.5653          & 0.8123          & 0.6393              & 0.6597          & 0.9688          & 0.8010          \\
CORE            & \textbf{1.0000} & 0.9998           & 0.8330          & 0.9960          & 0.8189              & 0.8272          & 0.9975          & 0.9246          \\
SPSL            & 0.9998          & 0.9971           & 0.5762          & 0.8188          & 0.7562              & 0.7588          & 0.9924          & 0.8427          \\
UIA-ViT         & \textbf{1.0000} & 0.9993           & 0.6704          & 0.9538          & 0.8145              & 0.7873          & 0.9949          & 0.8886          \\
DIRE            & 0.9997          & 0.9997           & 0.8315          & 0.9854          & 0.8542              & 0.8362          & 0.9968          & 0.9291          \\
UnivFD          & 0.9988          & 0.9954           & 0.9203          & 0.9915          & 0.9275              & 0.9854          & 0.9685          & 0.9696          \\
AIDE            & 0.9998          & 0.9897           & 0.7298          & 0.9149          & 0.8706              & 0.8479          & 0.9715          & 0.9035          \\
NPR             & \ul{0.9999}  & 0.9980           & \ul{0.9427}  & \ul{0.9964}  & 0.9479              & 0.9593          & 0.9874          & 0.9759          \\
Effort          & \textbf{1.0000} & \textbf{1.0000}  & 0.9088          & 0.9896          & \textbf{0.9803}     & \ul{0.9971}  & \ul{0.9998}  & \ul{0.9822}  \\
TriDetect             & \textbf{1.0000} & \textbf{1.0000}  & \textbf{0.9656} & \textbf{0.9992} & \ul{0.9616}      & \textbf{0.9992} & \textbf{0.9999} & \textbf{0.9894} 
\end{tblr}
\end{table}

% \definecolor{FrenchPass}{rgb}{0.741,0.878,0.996}
\begin{table}[ht]
\centering
\fontsize{6pt}{6pt}\selectfont
\caption{Comparison on the AIGCDetectBenchmark \cite{zhong2023patchcraft} Dataset in terms of ACC Performance. The best result and the second-best result are marked in \textbf{bold} and \ul{underline}, respectively.}
\label{table:AIGC-ACC}
\begin{tblr}{
  width = \linewidth,
  colspec = {Q[l, 80] *{17}{Q[c, 52]}}, % Centered columns
  cells = {c},
  row{1} = {cmd=\rotatebox{65}}, % Apply rotation to all cells in the first row
  row{15} = {FrenchPass},
  vline{2-18} = {-}{},
  hline{1,16} = {-}{0.08em},
  hline{2,15} = {-}{},
  hline{2} = {2}{-}{},
}
\textbf{Mehtod} & \textbf{CycleGAN} & \textbf{StyleGAN} & \textbf{StyleGAN2} & \textbf{ProGAN} & \textbf{GauGAN} & \textbf{StarGAN} & \textbf{BigGAN} & \textbf{ADM}    & \textbf{Wukong} & \textbf{Glide}  & \textbf{SD\_XL} & \textbf{VQDM}   & \textbf{MidJourney} & \textbf{SD v1.4} & \textbf{SD v1.5} & \textbf{DALLE2} & \textbf{Avg}    \\
CNNSpot         & 0.4974            & 0.5102            & 0.5029             & 0.4975          & 0.5027          & 0.5000           & 0.4858          & 0.5170          & 0.9658          & 0.5882          & 0.5200          & 0.5106          & 0.5344              & 0.9910           & 0.9903           & 0.5810          & 0.6059          \\
FreDect         & 0.5049            & 0.5645            & 0.5538             & 0.5405          & 0.5330          & 0.5720           & 0.7083          & 0.5368          & 0.9443          & 0.5533          & 0.6273          & 0.6113          & 0.5473              & 0.9720           & 0.9723           & 0.5530          & 0.6434          \\
Fusing          & 0.4890            & 0.5034            & 0.4991             & 0.5013          & 0.5022          & 0.4950           & 0.4978          & 0.5081          & 0.9718          & 0.5664          & 0.5163          & 0.5318          & 0.5098              & 0.9948           & 0.9928           & 0.5310          & 0.6007          \\
LGrad           & 0.4837            & 0.4819            & 0.4383             & 0.4970          & 0.4644          & 0.4985           & 0.4985          & 0.5169          & 0.5593          & 0.6264          & 0.4643          & 0.5078          & 0.5242              & 0.5346           & 0.5347           & 0.5885          & 0.5137          \\
LNP             & 0.4720            & 0.5000            & 0.4955             & 0.5000          & 0.4973          & 0.4855           & 0.4928          & 0.5059          & 0.8882          & 0.6103          & 0.5385          & 0.5398          & 0.5366              & 0.9177           & 0.9176           & 0.5570          & 0.5909          \\
CORE            & 0.5061            & 0.5094            & 0.5003             & 0.5083          & 0.5027          & 0.6153           & 0.5063          & 0.5684          & 0.9643          & 0.9498          & 0.5093          & 0.5581          & 0.5298              & 0.9933           & 0.9921           & 0.5920          & 0.6441          \\
SPSL            & 0.5091            & 0.4984            & 0.4970             & 0.5025          & 0.4999          & 0.4957           & 0.5093          & 0.5008          & 0.8925          & 0.5358          & 0.5180          & 0.5192          & 0.5292              & 0.9313           & 0.9296           & 0.5275          & 0.5872          \\
UIA-ViT         & 0.5220            & 0.5187            & 0.5225             & 0.4974          & 0.4989          & 0.5230           & 0.5228          & 0.5038          & 0.9322          & 0.7339          & 0.6313          & 0.5452          & 0.5768              & 0.9860           & 0.9851           & 0.5655          & 0.6291          \\
DIRE            & 0.5140            & 0.5495            & 0.5597             & 0.6496          & 0.4944          & 0.6601           & 0.5330          & 0.5985          & 0.9713          & 0.8973          & 0.6170          & 0.6055          & 0.6320              & \ul{0.9943}   & 0.9919           & 0.7605          & 0.6893          \\
UnivFD          & 0.8812            & 0.6551            & 0.6696             & 0.7349          & 0.9359          & 0.7691           & 0.8273          & 0.7212          & 0.8469          & 0.9482          & 0.8145          & 0.9143          & \ul{0.7583}      & 0.9653           & 0.9650           & 0.8715          & 0.8299          \\
AIDE            & 0.5473            & 0.6143            & 0.5933             & 0.7096          & 0.6633          & 0.5768           & 0.6613          & 0.5637          & 0.8972          & 0.7567          & 0.7510          & 0.6528          & 0.7174              & 0.9508           & 0.9499           & 0.7190          & 0.7078          \\
NPR             & 0.7354            & 0.8438            & \textbf{0.8496}    & 0.8986          & 0.5058          & \textbf{0.9927}  & 0.6993          & \textbf{0.7745} & 0.9177          & \textbf{0.9751} & \ul{0.8600}  & 0.8078          & \textbf{0.7748}     & 0.9792           & 0.9774           & \textbf{0.9635} & 0.8472          \\
Effort          & \ul{0.9387}    & \ul{0.8625}    & \ul{0.8201}     & \ul{0.9020}  & \textbf{0.9874} & \ul{0.9250}   & \textbf{0.9863} & 0.5872          & \ul{0.9878}  & 0.7942          & \textbf{0.8838} & \ul{0.9194}  & 0.7411              & \textbf{0.9993}  & \ul{0.9985}   & 0.7525          & \ul{0.8804}  \\
TriDetect             & \textbf{0.9974}   & \textbf{0.8654}   & 0.7702             & \textbf{0.9909} & \ul{0.9807}  & 0.9230           & \ul{0.9760}  & \ul{0.7482}  & \textbf{0.9963} & \ul{0.9488}  & 0.7915          & \textbf{0.9679} & 0.7480              & \textbf{0.9993}  & \textbf{0.9986}  & \ul{0.9405}  & \textbf{0.9152} 
\end{tblr}
\end{table}

\begin{table}[ht]
\centering
\fontsize{6pt}{6pt}\selectfont
\caption{Comparison on the AIGCDetectBenchmark \cite{zhong2023patchcraft} Dataset in terms of EER Performance. The best result and the second-best result are marked in \textbf{bold} and \ul{underline}, respectively.}
\label{table:AIGC-EER}
\begin{tblr}{
  width = \linewidth,
  colspec = {Q[l, 80] *{17}{Q[c, 52]}}, % Centered columns
  cells = {c},
  row{1} = {cmd=\rotatebox{65}}, % Apply rotation to all cells in the first row
  row{15} = {FrenchPass},
  vline{2-18} = {-}{},
  hline{1,16} = {-}{0.08em},
  hline{2,15} = {-}{},
  hline{2} = {2}{-}{},
}
\textbf{Mehtod} & \textbf{CycleGAN} & \textbf{StyleGAN} & \textbf{StyleGAN2} & \textbf{ProGAN} & \textbf{GauGAN} & \textbf{StarGAN} & \textbf{BigGAN} & \textbf{ADM}    & \textbf{Wukong} & \textbf{Glide}          & \textbf{SD\_XL} & \textbf{VQDM}   & \textbf{MidJourney} & \textbf{SD v1.4} & \textbf{SD v1.5} & \textbf{DALLE2} & \textbf{Avg}    \\
CNNSpot         & 0.6109            & 0.4782            & 0.5159             & 0.4923          & 0.5556          & 0.4562           & 0.5760          & 0.2595          & 0.0153          & 0.1048                  & 0.2555          & 0.2747          & 0.2815              & 0.0052           & 0.0063           & 0.0740          & 0.3101          \\
FreDect         & 0.5783            & 0.3893            & 0.3708             & 0.3425          & 0.4466          & 0.3382           & 0.2615          & 0.3145          & 0.0525          & 0.3017                  & 0.2040          & 0.2697          & 0.3275              & 0.0275           & 0.0279           & 0.2530          & 0.2816          \\
Fusing          & 0.5950            & 0.4877            & 0.4735             & 0.5190          & 0.5150          & 0.4912           & 0.5925          & 0.3150          & \ul{0.0130}  & 0.1185                  & 0.2925          & 0.2898          & 0.3637              & \ul{0.0045}   & 0.0063           & 0.0970          & 0.3234          \\
LGrad           & 0.5185            & 0.5196            & 0.5628             & 0.4998          & 0.5326          & 0.5063           & 0.5040          & 0.4830          & 0.4373          & 0.3683                  & 0.5375          & 0.4912          & 0.4705              & 0.4642           & 0.4614           & 0.4130          & 0.5909          \\
LNP             & 0.5731            & 0.4921            & 0.5091             & 0.5008          & 0.5068          & 0.5528           & 0.5490          & 0.4257          & 0.0920          & 0.2358                  & 0.3415          & 0.3695          & 0.3880              & 0.0702           & 0.0713           & 0.2470          & 0.3703          \\
CORE            & 0.4232            & 0.4260            & 0.4384             & 0.4560          & 0.4814          & 0.0860           & 0.4590          & 0.2562          & 0.0203          & 0.0273                  & 0.3045          & 0.2570          & 0.2627              & 0.0050           & 0.0066           & 0.1070          & 0.2510          \\
SPSL            & 0.4391            & 0.4754            & 0.4772             & 0.4813          & 0.4840          & 0.4782           & 0.3905          & 0.4323          & 0.0430          & 0.2482                  & 0.2485          & 0.2935          & 0.3167              & 0.0267           & 0.0298           & 0.2370          & 0.3188          \\
UIA-ViT         & 0.4981            & 0.3475            & 0.3065             & 0.4970          & 0.5086          & 0.4957           & 0.3905          & 0.3313          & 0.0348          & 0.1227                  & 0.1375          & 0.2822          & 0.2813              & 0.0117           & 0.0134           & 0.2610          & 0.2825          \\
DIRE            & 0.4164            & 0.4225            & 0.4007             & 0.2790          & 0.5722          & 0.0360           & 0.4655          & 0.2517          & 0.0255          & 0.0632                  & 0.1890          & 0.2540          & 0.2383              & 0.0053           & 0.0075           & 0.1010          & 0.2330          \\
UnivFD          & 0.0984            & 0.3402            & 0.3294             & 0.1270          & 0.0630          & 0.1536           & 0.0895          & 0.1527          & 0.0922          & 0.0433                  & 0.1045          & 0.0593          & 0.1528              & 0.0317           & 0.0324           & 0.0970          & 0.1229          \\
AIDE            & 0.4164            & 0.3372            & 0.3400             & 0.2735          & 0.2330          & 0.3507           & 0.2575          & 0.2968          & 0.0882          & 0.1463                  & 0.1425          & 0.2067          & 0.2127              & 0.0493           & 0.0491           & 0.1880          & 0.2242          \\
NPR             & 0.1499            & 0.1751            & \ul{0.1383}     & 0.1028          & 0.4756          & \ul{0.0040}   & 0.2865          & 0.1462          & 0.0610          & \textbf{\ul{0.0260}} & 0.0840          & 0.1105          & 0.1302              & 0.0183           & 0.0216           & \ul{0.0370}  & 0.1229          \\
Effort          & \ul{0.0507}    & \ul{0.1349}    & 0.1852             & \ul{0.0238}  & \textbf{0.0090} & 0.0390           & \ul{0.0130}  & \ul{0.1763}  & \ul{0.0067}  & 0.0520                  & \textbf{0.0215} & \ul{0.0253}  & \textbf{0.0768}     & \textbf{0.0007}  & \ul{0.0015}   & 0.0960          & \ul{0.0570}  \\
TriDetect             & \textbf{0.0015}   & \textbf{0.1249}   & \textbf{0.1253}    & \textbf{0.0048} & \ul{0.0220}  & \textbf{0.0000}  & \textbf{0.0120} & \textbf{0.1032} & \textbf{0.0030} & \textbf{0.0087}         & \ul{0.0325}  & \textbf{0.0110} & \ul{0.1133}      & \textbf{0.0007}  & \textbf{0.0001}  & \textbf{0.0240} & \textbf{0.0367} 
\end{tblr}
\end{table}

\begin{table}[ht]
\centering
\fontsize{6pt}{6pt}\selectfont
\caption{Comparison on the AIGCDetectBenchmark \cite{zhong2023patchcraft} Dataset in terms of AP Performance. The best result and the second-best result are marked in \textbf{bold} and \ul{underline}, respectively.}
\label{table:AIGC-AP}
\begin{tblr}{
  width = \linewidth,
  colspec = {Q[l, 80] *{17}{Q[c, 52]}}, % Centered columns
  cells = {c},
  row{1} = {cmd=\rotatebox{65}}, % Apply rotation to all cells in the first row
  row{15} = {FrenchPass},
  vline{2-18} = {-}{},
  hline{1,16} = {-}{0.08em},
  hline{2,15} = {-}{},
  hline{2} = {2}{-}{},
}
\textbf{Mehtod} & \textbf{CycleGAN} & \textbf{StyleGAN} & \textbf{StyleGAN2} & \textbf{ProGAN} & \textbf{GauGAN} & \textbf{StarGAN} & \textbf{BigGAN} & \textbf{ADM}    & \textbf{Wukong} & \textbf{Glide}  & \textbf{SD\_XL} & \textbf{VQDM}   & \textbf{MidJourney} & \textbf{SD v1.4} & \textbf{SD v1.5} & \textbf{DALLE2} & \textbf{Avg}    \\
CNNSpot         & 0.4101            & 0.5390            & 0.4980             & 0.5021          & 0.4636          & 0.5383           & 0.4183          & 0.8075          & 0.9989          & 0.9518          & 0.8329          & 0.7847          & 0.8025              & \ul{0.9999}   & 0.9995           & 0.9701          & 0.7198          \\
FreDect         & 0.4631            & 0.6684            & 0.6745             & 0.7120          & 0.5859          & 0.7127           & 0.8452          & 0.7140          & 0.9902          & 0.7391          & 0.8651          & 0.8049          & 0.7193              & 0.9971           & 0.9961           & 0.8070          & 0.7684          \\
Fusing          & 0.4266            & 0.5229            & 0.5176             & 0.4896          & 0.5019          & 0.4814           & 0.4296          & 0.7288          & 0.9987          & 0.9332          & 0.7774          & 0.7767          & 0.6845              & \ul{0.9999}   & \ul{0.9998}   & 0.9465          & 0.7009          \\
LGrad           & 0.4901            & 0.4992            & 0.4517             & 0.5109          & 0.4777          & 0.5089           & 0.5090          & 0.5226          & 0.6290          & 0.6321          & 0.5079          & 0.5037          & 0.5306              & 0.5937           & 0.5924           & 0.5611          & 0.5325          \\
LNP             & 0.4154            & 0.5083            & 0.4805             & 0.5012          & 0.4841          & 0.4459           & 0.4621          & 0.5652          & 0.9688          & 0.8129          & 0.6998          & 0.6597          & 0.6395              & 0.9818           & 0.9801           & 0.8051          & 0.6507          \\
CORE            & 0.5821            & 0.6144            & 0.5383             & 0.5614          & 0.5225          & 0.9743           & 0.5522          & 0.8330          & 0.9975          & 0.9960          & 0.7440          & 0.8272          & 0.8189              & 0.9998           & 0.9997           & 0.9492          & 0.7819          \\
SPSL            & 0.5695            & 0.5145            & 0.5038             & 0.5201          & 0.5029          & 0.4950           & 0.6142          & 0.5762          & 0.9924          & 0.8188          & 0.8146          & 0.7588          & 0.7562              & 0.9971           & 0.9957           & 0.8357          & 0.7041          \\
UIA-ViT         & 0.5417            & 0.6884            & 0.7113             & 0.5022          & 0.4890          & 0.5831           & 0.6207          & 0.6704          & 0.9949          & 0.9538          & 0.9242          & 0.7873          & 0.8145              & 0.9993           & 0.9988           & 0.8045          & 0.7553          \\
DIRE            & 0.5985            & 0.6383            & 0.6571             & 0.7953          & 0.4323          & 0.9937           & 0.5858          & 0.8315          & 0.9968          & 0.9854          & 0.8875          & 0.8362          & 0.8542              & 0.9997           & 0.9993           & 0.9620          & 0.8159          \\
UnivFD          & 0.9641            & 0.7507            & 0.7300             & 0.9301          & 0.9798          & 0.8900           & 0.9704          & 0.9203          & 0.9685          & 0.9915          & 0.9598          & 0.9854          & 0.9275              & 0.9954           & 0.9951           & 0.9562          & 0.9322          \\
AIDE            & 0.5267            & 0.7119            & 0.6912             & 0.7424          & 0.8059          & 0.6882           & 0.7557          & 0.7298          & 0.9715          & 0.9149          & 0.9246          & 0.8479          & 0.8706              & 0.9897           & 0.9901           & 0.8723          & 0.8146          \\
NPR             & 0.9127            & 0.9208            & \textbf{0.9486}    & 0.9681          & 0.5425          & \ul{0.9998}   & 0.7852          & \ul{0.9427}  & 0.9874          & \ul{0.9964}  & 0.9757          & 0.9593          & 0.9479              & 0.9980           & 0.9973           & 0.9945          & 0.9298          \\
Effort          & \ul{0.9899}    & \ul{0.9510}    & 0.9151             & \ul{0.9973}  & \textbf{0.9997} & 0.9919           & \ul{0.9991}  & 0.9088          & \ul{0.9998}  & 0.9896          & \textbf{0.9979} & \ul{0.9971}  & \textbf{0.9803}     & \textbf{1.0000}  & \textbf{1.0000}  & \ul{0.9647}  & \ul{0.9801}  \\
TriDetect             & \textbf{1.0000}   & \textbf{0.9575}   & 0.9020             & \textbf{0.9996} & \ul{0.9976}  & \textbf{1.0000}  & \textbf{0.9992} & \textbf{0.9656} & \textbf{0.9999} & \textbf{0.9992} & \ul{0.9940}  & \textbf{0.9992} & \ul{0.9616}      & \textbf{1.0000}  & \textbf{1.0000}  & \textbf{0.9969} & \textbf{0.9858} 
\end{tblr}
\end{table}

\begin{table*}[ht]
\centering
\fontsize{7pt}{7pt}\selectfont
\caption{Comparison on the WildFake \cite{hong2025wildfake} Dataset in terms of AUC Performance. The best result and the second-best result are marked in \textbf{bold} and \ul{underline}, respectively.}
\label{table:WildFake-AUC}
\begin{tblr}{
  width = \linewidth,
  colspec = {Q[85]Q[77]Q[67]Q[67]Q[67]Q[79]Q[85]Q[90]Q[81]Q[67]Q[88]Q[67]},
  cells = {c},
  row{15} = {FrenchPass},
  vline{2-12} = {-}{},
  hline{1,16} = {-}{0.08em},
  hline{2,15} = {-}{},
  hline{2} = {2}{-}{},
}
\textbf{Method} & \textbf{DALL-E} & \textbf{DDIM}   & \textbf{DDPM}   & \textbf{VQDM}   & \textbf{BigGAN} & \textbf{StarGAN} & \textbf{StyleGAN} & \textbf{DF-GAN} & \textbf{GALIP}  & \textbf{GigaGAN} & \textbf{Avg}    \\
CNNSpot         & 0.8220          & 0.5943          & 0.3375          & 0.3706          & 0.9513          & 0.5003           & 0.4613            & 0.5304          & 0.5143          & 0.4285           & 0.5511          \\
FreDect         & 0.7546          & 0.7250          & 0.7385          & 0.8559          & \ul{0.9867}  & 0.7985           & 0.5242            & 0.8151          & 0.7686          & 0.8864           & 0.7854          \\
Fusing          & 0.8371          & 0.7024          & 0.4468          & 0.5436          & \textbf{0.9972} & 0.8057           & 0.5858            & 0.5975          & 0.8207          & 0.4651           & 0.6802          \\
LGrad           & 0.6232          & 0.5193          & 0.5236          & 0.5065          & 0.5593          & 0.5172           & 0.4637            & 0.5482          & 0.4033          & 0.5782           & 0.5243          \\
LNP             & 0.8371          & 0.6025          & 0.4387          & 0.6140          & 0.9401          & 0.7557           & 0.6193            & 0.6148          & 0.6199          & 0.5675           & 0.6610          \\
CORE            & \textbf{0.9213} & 0.7088          & 0.5795          & 0.8723          & 0.9224          & 0.7298           & 0.5879            & 0.9104          & 0.7876          & 0.7192           & 0.7739          \\
SPSL            & 0.8470          & 0.3266          & 0.2551          & 0.5248          & 0.9787          & 0.4803           & 0.5476            & 0.5860          & 0.4299          & 0.2803           & 0.5256          \\
UIA-ViT         & 0.9045          & 0.4139          & 0.3342          & 0.6346          & 0.8994          & 0.5121           & 0.5507            & 0.7466          & 0.6520          & 0.5660           & 0.6214          \\
DIRE            & 0.8880          & 0.4755          & 0.4363          & 0.8621          & 0.6951          & 0.5949           & 0.5216            & 0.8632          & 0.8636          & 0.7628           & 0.6963          \\
UnivFD          & 0.5857          & 0.8008          & \ul{0.7873}  & 0.7802          & 0.8711          & 0.8779           & 0.6156            & 0.9597          & 0.9257          & 0.8417           & 0.8046          \\
AIDE            & 0.6062          & 0.4235          & 0.4369          & 0.6714          & 0.8322          & 0.3749           & 0.5730            & 0.8317          & 0.5975          & 0.3711           & 0.5718          \\
NPR             & 0.8056          & \textbf{0.9063} & \textbf{0.7906} & 0.9339          & 0.9128          & 0.8022           & 0.5161            & 0.9233          & 0.7018          & 0.8276           & 0.8120          \\
Effort          & 0.8537          & 0.8486          & 0.7197          & \ul{0.9372}  & 0.9599          & \ul{0.9023}   & \textbf{0.7605}   & \ul{0.9994}  & \ul{0.9343}  & \ul{0.9013}   & \ul{0.8817}  \\
TriDetect            & \ul{0.9189}  & \ul{0.8823}  & 0.7106          & \textbf{0.9787} & 0.9802          & \textbf{1.0000}  & \ul{0.7245}    & \textbf{1.0000} & \textbf{0.9774} & \textbf{0.9981}  & \textbf{0.9171}
\end{tblr}
\end{table*}

\begin{table}[ht]
\centering
\fontsize{7pt}{7pt}\selectfont
\caption{Comparison on the WildFake \cite{hong2025wildfake} Dataset in terms of EER Performance. The best result and the second-best result are marked in \textbf{bold} and \ul{underline}, respectively.}
\label{table:wildfake-EER}
\begin{tblr}{
  width = \linewidth,
  colspec = {Q[87]Q[77]Q[67]Q[67]Q[67]Q[81]Q[85]Q[92]Q[81]Q[67]Q[88]Q[67]},
  cells = {c},
  row{15} = {FrenchPass},
  vline{2-12} = {-}{},
  hline{1,16} = {-}{0.08em},
  hline{2,15} = {-}{},
  hline{2} = {2}{-}{},
}
\textbf{Method} & \textbf{DALL-E} & \textbf{DDIM}   & \textbf{DDPM}   & \textbf{VQDM}   & \textbf{BigGAN} & \textbf{StarGAN} & \textbf{StyleGAN} & \textbf{DF-GAN} & \textbf{GALIP}  & \textbf{GigaGAN} & \textbf{Avg}    \\
CNNSpot         & 0.2296          & 0.4286          & 0.6301          & 0.5887          & 0.1236          & 0.5006           & 0.5271            & 0.4693          & 0.4796          & 0.5418           & 0.4519          \\
FreDect         & 0.2302          & 0.2561          & \ul{0.3288}  & 0.2301          & 0.0650          & 0.2778           & 0.4790            & 0.2318          & 0.3013          & \ul{0.0588}   & 0.2459          \\
Fusing          & 0.2215          & 0.3521          & 0.5538          & 0.4813          & \textbf{0.0277} & 0.2662           & 0.4341            & 0.4402          & 0.2556          & 0.5452           & 0.3578          \\
LGrad           & 0.4117          & 0.4818          & 0.4721          & 0.4953          & 0.4582          & 0.4896           & 0.5266            & 0.4692          & 0.5731          & 0.4405           & 0.4818          \\
LNP             & 0.2265          & 0.4292          & 0.5489          & 0.4160          & 0.1371          & 0.2999           & 0.4151            & 0.4153          & 0.4177          & 0.4476           & 0.3753          \\
CORE            & 0.1409          & 0.3562          & 0.4678          & 0.2096          & 0.1274          & 0.3135           & 0.4390            & 0.1658          & 0.2815          & 0.3399           & 0.2842          \\
SPSL            & 0.2116          & 0.6288          & 0.6992          & 0.4826          & 0.0714          & 0.5084           & 0.4611            & 0.4417          & 0.5530          & 0.6577           & 0.4716          \\
UIA-ViT         & \ul{0.1731}  & 0.5870          & 0.6240          & 0.4113          & 0.1885          & 0.5052           & 0.4705            & 0.3229          & 0.3966          & 0.4452           & 0.4124          \\
DIRE            & 0.1939          & 0.5424          & 0.5703          & 0.2235          & 0.3649          & 0.4398           & 0.4888            & 0.2149          & 0.2101          & 0.3060           & 0.3554          \\
UnivFD          & 0.4365          & 0.2179          & 0.2803          & 0.2327          & 0.2162          & \ul{0.1839}   & \ul{0.4214}    & 0.0445          & \ul{0.1504}  & 0.2397           & 0.2423          \\
AIDE            & 0.4254          & 0.5607          & 0.5601          & 0.3741          & 0.2535          & 0.5835           & 0.4479            & 0.2408          & 0.4244          & 0.5906           & 0.4461          \\
NPR             & 0.2507          & \textbf{0.1748} & \textbf{0.2731} & \ul{0.1376}  & 0.1847          & 0.2895           & 0.4903            & 0.1159          & 0.3496          & 0.1906           & 0.2457          \\
Effort          & 0.2268          & 0.2376          & 0.3516          & 0.1394          & 0.1055          & 0.1859           & \textbf{0.3091}   & \ul{0.0106}  & 0.1505          & 0.1795           & \ul{0.1897}  \\
TriDetect             & \textbf{0.1508} & \ul{0.1898}  & 0.3415          & \textbf{0.0691} & \ul{0.0663}  & \textbf{0.0000}  & \textbf{0.3391}   & \textbf{0.0020} & \textbf{0.0866} & \textbf{0.0167}  & \textbf{0.1262} 
\end{tblr}
\end{table}

\begin{table}[ht]
\fontsize{7pt}{7pt}\selectfont
\caption{Comparison on the WildFake \cite{hong2025wildfake} Dataset in terms of AP Performance. The best result and the second-best result are marked in \textbf{bold} and \ul{underline}, respectively.}
\label{table:wildfake-AP}
\begin{tblr}{
  width = \linewidth,
  colspec = {Q[87]Q[77]Q[67]Q[67]Q[67]Q[81]Q[85]Q[92]Q[81]Q[67]Q[88]Q[67]},
  cells = {c},
  row{15} = {FrenchPass},
  vline{2-12} = {-}{},
  hline{1,16} = {-}{0.08em},
  hline{2,15} = {-}{},
  hline{2} = {2}{-}{},
}
\textbf{Method} & \textbf{DALL-E} & \textbf{DDIM}   & \textbf{DDPM}   & \textbf{VQDM}   & \textbf{BigGAN} & \textbf{StarGAN} & \textbf{StyleGAN} & \textbf{DF-GAN} & \textbf{GALIP}  & \textbf{GigaGAN} & \textbf{Avg}    \\
CNNSpot         & 0.9519          & 0.8544          & 0.7365          & 0.6506          & 0.9776          & 0.6501           & 0.6442            & 0.6866          & 0.6657          & 0.6288           & 0.7447          \\
FreDect         & 0.9604          & 0.9448          & 0.9048          & 0.9369          & \ul{0.9936}  & 0.8776           & 0.6762            & 0.9163          & 0.8374          & \ul{0.9934}   & 0.9041          \\
Fusing          & 0.9592          & 0.9008          & 0.7733          & 0.7869          & \textbf{0.9986} & 0.8655           & 0.7226            & 0.7607          & 0.8838          & 0.6746           & 0.8326          \\
LGrad           & 0.8769          & 0.8221          & 0.8005          & 0.7408          & 0.7095          & 0.6816           & 0.6470            & 0.7226          & 0.6294          & 0.7465           & 0.7377          \\
LNP             & 0.9544          & 0.8586          & 0.7782          & 0.8167          & 0.9679          & 0.8244           & 0.7607            & 0.7595          & 0.7562          & 0.7309           & 0.8207          \\
CORE            & \textbf{0.9801} & 0.9115          & 0.8480          & 0.9419          & 0.9681          & 0.8426           & 0.7368            & 0.9502          & 0.8608          & 0.8340           & 0.8874          \\
SPSL            & 0.9591          & 0.7492          & 0.6971          & 0.7532          & 0.9894          & 0.6402           & 0.6913            & 0.7486          & 0.6263          & 0.5519           & 0.7406          \\
UIA-ViT         & 0.9730          & 0.7965          & 0.7393          & 0.8210          & 0.9537          & 0.6873           & 0.7224            & 0.8444          & 0.7614          & 0.7281           & 0.8027          \\
DIRE            & 0.9694          & 0.8281          & 0.7883          & 0.9326          & 0.8625          & 0.7693           & 0.6945            & 0.9076          & 0.8957          & 0.8312           & 0.8479          \\
UnivFD          & 0.8649          & 0.9606          & \ul{0.9280}  & 0.9304          & 0.9345          & 0.9464           & 0.7535            & 0.9937          & 0.9601          & 0.9204           & \ul{0.9192}  \\
AIDE            & 0.8525          & 0.7664          & 0.7549          & 0.7987          & 0.9202          & 0.5747           & 0.6885            & 0.8673          & 0.7055          & 0.5808           & 0.7510          \\
NPR             & 0.9452          & \textbf{0.9748} & \textbf{0.9356} & \ul{0.9745}  & 0.9545          & 0.8888           & 0.6931            & 0.9775          & 0.8554          & 0.9435           & 0.9143          \\
Effort          & 0.9532          & 0.9490          & 0.8936          & 0.9744          & 0.9807          & \ul{0.9493}   & \textbf{0.8589}   & \ul{0.9997}  & \ul{0.9652}  & 0.9412           & \ul{0.9465}  \\
TriDetect             & \ul{0.9794}  & \ul{0.9638}  & 0.8922          & \textbf{0.9924} & 0.9914          & \textbf{1.0000}  & \ul{0.8385}    & \textbf{1.0000} & \textbf{0.9887} & \textbf{0.9991}  & \textbf{0.9645} 
\end{tblr}
\end{table}

\begin{table*}[ht]
\centering
\fontsize{7pt}{7pt}\selectfont
\caption{Comparison on the DF40 \cite{yan2024df40} Dataset in terms of AUC Performance. The best result and the second-best result are marked in \textbf{bold} and \ul{underline}, respectively.}
\label{table:DF40-AUC}
\begin{tblr}{
  width = \linewidth,
  colspec = {Q[130]Q[102]Q[117]Q[100]Q[92]Q[102]Q[100]Q[100]Q[73]},
  cells = {c},
  row{15} = {FrenchPass},
  vline{2-9} = {-}{},
  hline{1,16} = {-}{0.08em},
  hline{2,15} = {-}{},
  hline{2} = {2}{-}{},
}
\textbf{Method} & \textbf{CollabDiff} & \textbf{MidJourney} & \textbf{DeepFaceLab} & \textbf{StarGAN} & \textbf{StarGAN2} & \textbf{StyleCLIP} & \textbf{WhichisReal} & \textbf{Avg}    \\
CNNSpot         & 0.8890              & 0.3994              & 0.5113               & 0.3954           & 0.4860            & 0.7803             & 0.6265               & 0.5840          \\
FreDect         & 0.9764              & 0.3734              & 0.5534               & 0.5091           & 0.4825            & 0.8425             & 0.5550               & 0.6132          \\
Fusing          & 0.8969              & 0.3569              & \ul{0.6875}       & 0.3920           & 0.5091            & 0.6957             & 0.6581               & 0.5994          \\
LGrad           & 0.5063              & 0.4947              & 0.4998               & 0.5078           & 0.4508            & 0.4746             & 0.5185               & 0.4932          \\
LNP             & 0.5320              & 0.3744              & 0.5476               & 0.4602           & 0.5256            & 0.6324             & 0.6144               & 0.5266          \\
CORE            & 0.8793              & 0.4534              & 0.6239               & 0.5729           & 0.4340            & 0.7097             & 0.3048               & 0.5683          \\
SPSL            & 0.7371              & 0.4621              & 0.5203               & 0.5078           & 0.5399            & 0.7620             & 0.6220               & 0.5930          \\
UIA-ViT         & 0.9334              & 0.2430              & 0.5255               & 0.3849           & 0.5068            & 0.9731             & 0.4437               & 0.5729          \\
DIRE            & 0.9441              & 0.6177              & 0.3853               & 0.3796           & 0.4355            & 0.8546             & 0.4139               & 0.5758          \\
UnivFD          & 0.6389              & 0.8767              & 0.5013               & 0.8650           & 0.7563            & 0.8841             & 0.7555               & 0.7540          \\
NPR             & 0.0536              & \textbf{0.9541}     & 0.3740               & 0.4929           & 0.4542            & 0.0409             & 0.8535               & 0.4605          \\
AIDE            & 0.7235              & 0.8004              & 0.3989               & 0.7596           & 0.5697            & 0.7630             & 0.6294               & 0.6635          \\
Effort          & \textbf{0.9991}     & \ul{0.9276}      & 0.5496               & \ul{0.9859}   & \ul{0.9349}    & \textbf{0.9993}    & \textbf{0.9122}      & \ul{0.9012}  \\
TriDetect             & \ul{0.9967}      & 0.9147              & \textbf{0.7493}      & \textbf{0.9984}  & \textbf{0.9450}   & \ul{0.9986}     & \ul{0.9080}       & \textbf{0.9301}
\end{tblr}
\end{table*}

\begin{table}[ht]
\centering
\fontsize{7pt}{7pt}\selectfont
\caption{Comparison on the DF40 \cite{yan2024df40} Dataset in terms of EER Performance. The best result and the second-best result are marked in \textbf{bold} and \ul{underline}, respectively.}
\label{table:DF40-EER}
\begin{tblr}{
  width = \linewidth,
  colspec = {Q[130]Q[102]Q[117]Q[100]Q[92]Q[102]Q[100]Q[100]Q[73]},
  cells = {c},
  row{15} = {FrenchPass},
  vline{2-9} = {-}{},
  hline{1,16} = {-}{0.08em},
  hline{2,15} = {-}{},
  hline{2} = {2}{-}{},
}
\textbf{Method} & \textbf{CollabDiff} & \textbf{MidJourney} & \textbf{DeepFaceLab} & \textbf{StarGAN} & \textbf{StarGAN2} & \textbf{StyleCLIP} & \textbf{WhichisReal} & \textbf{Avg}    \\
CNNSpot         & 0.1920              & 0.5660              & 0.4984               & 0.5766           & 0.5035            & 0.2800             & 0.3980               & 0.4306          \\
FreDect         & 0.0850              & 0.6010              & 0.4609               & 0.4950           & 0.5060            & 0.2520             & 0.4700               & 0.4100          \\
Fusing          & 0.1800              & 0.6130              & \ul{0.3601}       & 0.5806           & 0.4920            & 0.3500             & 0.3750               & 0.4215          \\
LGrad           & 0.4960              & 0.4980              & 0.4990               & 0.4990           & 0.5320            & 0.5290             & 0.4870               & 0.5057          \\
LNP             & 0.4770              & 0.5890              & 0.4628               & 0.5232           & 0.4865            & 0.4120             & 0.4200               & 0.4815          \\
CORE            & 0.1950              & 0.5500              & 0.4066               & 0.4410           & 0.5360            & 0.3360             & 0.6340               & 0.4427          \\
SPSL            & 0.3230              & 0.5300              & 0.4971               & 0.4894           & 0.4715            & 0.3110             & 0.4090               & 0.4330          \\
UIA-ViT         & 0.1390              & 0.6950              & 0.4699               & 0.5882           & 0.5035            & \textbf{0.0820}    & 0.5440               & 0.4317          \\
DIRE            & 0.1320              & 0.4210              & 0.5779               & 0.5781           & 0.5415            & 0.2340             & 0.5420               & 0.4324          \\
UnivFD          & 0.4070              & 0.2020              & 0.5288               & 0.2188           & 0.3123            & 0.1990             & 0.3110               & 0.3113          \\
NPR             & 0.8990              & \ul{0.1150}      & 0.5928               & 0.5045           & 0.5370            & 0.8870             & 0.2250               & 0.5372          \\
AIDE            & 0.3370              & 0.2760              & 0.5779               & 0.3029           & 0.4444            & 0.3080             & 0.4080               & 0.3792          \\
Effort          & \textbf{0.0110}     & 0.1490              & 0.4641               & \ul{0.0590}   & \ul{0.1381}    & \ul{0.0120}     & \ul{0.1770}       & \ul{0.1443}  \\
TriDetect         & \ul{0.0330}      & \textbf{0.0720}     & \textbf{0.3368}      & \textbf{0.0112}  & \textbf{0.1326}   & 0.0150             & \textbf{0.1700}      & \textbf{0.1229} 
\end{tblr}
\end{table}

\begin{table}[ht]
\centering
\fontsize{7pt}{7pt}\selectfont
\caption{Comparison on the DF40 \cite{yan2024df40} Dataset in terms of AP Performance. The best result and the second-best result are marked in \textbf{bold} and \ul{underline}, respectively.}
\label{table:DF40-AP}
\begin{tblr}{
  width = \linewidth,
  colspec = {Q[130]Q[102]Q[117]Q[100]Q[92]Q[102]Q[100]Q[100]Q[73]},
  cells = {c},
  row{15} = {FrenchPass},
  vline{2-9} = {-}{},
  hline{1,16} = {-}{0.08em},
  hline{2,15} = {-}{},
  hline{2} = {2}{-}{},
}
\textbf{Method} & \textbf{CollabDiff} & \textbf{MidJourney} & \textbf{DeepFaceLab} & \textbf{StarGAN} & \textbf{StarGAN2} & \textbf{StyleCLIP} & \textbf{WhichisReal} & \textbf{Avg}    \\
CNNSpot         & 0.8976              & 0.4415              & 0.6559               & 0.4226           & 0.4949            & 0.9150             & 0.5899               & 0.6310          \\
FreDect         & 0.9798              & 0.3749              & 0.6882               & 0.4961           & 0.5086            & 0.9343             & 0.5533               & 0.6479          \\
Fusing          & 0.9057              & 0.3842              & 0.7736               & 0.4139           & 0.5091            & 0.8759             & 0.6037               & 0.6380          \\
LGrad           & 0.5249              & 0.4650              & 0.6720               & 0.5044           & 0.4566            & 0.7427             & 0.5309               & 0.5567          \\
LNP             & 0.5385              & 0.3777              & 0.7004               & 0.4676           & 0.5304            & 0.8215             & 0.5973               & 0.5762          \\
CORE            & 0.8593              & 0.4463              & 0.7018               & 0.5536           & 0.4407            & 0.8331             & 0.3743               & 0.6013          \\
SPSL            & 0.7233              & 0.4231              & 0.6734               & 0.4813           & 0.5292            & 0.8822             & 0.5752               & 0.6125          \\
UIA-ViT         & 0.9303              & 0.3268              & 0.6933               & 0.4465           & 0.5091            & 0.9903             & 0.4389               & 0.6193          \\
DIRE            & 0.9382              & 0.5679              & 0.5921               & 0.4201           & 0.4487            & 0.9387             & 0.4266               & 0.6189          \\
UnivFD          & 0.6131              & 0.8210              & 0.6957               & 0.8439           & 0.7150            & 0.9539             & 0.7066               & 0.7642          \\
NPR             & 0.3158              & \textbf{0.9549}     & 0.5802               & 0.4862           & 0.4690            & 0.5305             & 0.8374               & 0.5963          \\
AIDE            & 0.6908              & 0.7314              & 0.5838               & 0.7634           & 0.5311            & 0.8809             & 0.5686               & 0.6786          \\
Effort          & \textbf{0.9992}     & \ul{0.9012}      & \ul{0.6950}       & \ul{0.9851}   & \ul{0.9337}    & \textbf{0.9998}    & \ul{0.9042}       & \ul{0.9169}  \\
TriDetect         & \ul{0.9967}      & 0.8635              & \textbf{0.8331}      & \textbf{0.9985}  & \textbf{0.9461}   & \ul{0.9995}     & \textbf{0.9018}      & \textbf{0.9342} 
\end{tblr}
\end{table}

\newpage
\section{Limitation and Future Work}

In our manuscript and supplementary, we have provided detailed proofs for our theoretical analysis and conducted comprehensive experimental results and ablation studies to validate our proposed method - TriDetect. Results show that by discovering latent clusters within synthetic images that correspond to different generative architectures, TriDetect can improve the generalization capabilities to unseen generators across 5 datasets. One limitation of our work is that our method can generalize to unseen generators within the same architecture families as GANs and DMs, potentially failing to detect synthetic images generated by other architectures (e.g., VAE, normalizing flows).

In the future, we plan to extend our method into \textit{open-set recognition} settings where the model can not only detect known architectural patterns but also identify and flag images from entirely novel generation paradigms. This would involve developing an outlier detection mechanism that monitors the distance between test samples and learned cluster centroids, enabling the system to recognize when an image exhibits patterns inconsistent with both real images and known synthetic architectures. Another promising direction is the development of \textit{adaptive clustering mechanisms} that can dynamically adjust the number of clusters based on the observed data distribution. 
% \input{AnonymousSubmission/LaTeX/ReproducibilityChecklist}

% % \onecolumn
% \appendix
\end{document}